\documentclass[times,twocolumn,final, authoryear]{elsarticle}
\usepackage{graphicx}
\usepackage{subcaption}
\usepackage[table]{xcolor}
\usepackage{colortbl}
\usepackage{medima}
\usepackage{framed,multirow}
\usepackage{enumitem}
\usepackage{subfiles} 
\usepackage{amssymb}
\usepackage{amsmath}
\usepackage{latexsym}
\usepackage{bm}
\usepackage{tabularx}
\usepackage{rotating}
\usepackage{array}
\usepackage{makecell}
\usepackage{url}
\usepackage{booktabs}
\usepackage{hyperref}
\usepackage{xurl}
\usepackage{pifont}
\newcommand{\cmark}{\ding{51}}
\newcommand{\xmark}{\ding{55}}

\definecolor{cebg}{RGB}{240,245,255}
\definecolor{ncbg}{RGB}{245,240,255}
\definecolor{rowlight}{gray}{0.97}
\definecolor{mygray}{RGB}{230,230,230}

\definecolor{hdrbg}{RGB}{242,243,246}
\definecolor{blkbg}{RGB}{249,249,251}
\providecommand{\tabci}[1]{{\scriptsize\color{black!60}#1}}
\providecommand{\refcell}{\textit{\color{black!55}ref.}}
\newcolumntype{?}{!{\color{black!25}\vrule width .4pt}}
\newsavebox{\lifstab}
\newlength{\lifstabw}


\newcommand{\tabnote}[1]{\par\vspace{0.35em}\setlength{\lifstabw}{\wd\lifstab}\ifdim\lifstabw<1pt\setlength{\lifstabw}{\linewidth}\fi
  \noindent\begin{minipage}{\lifstabw}\footnotesize\color{black!75}\raggedright #1\end{minipage}}

\newcommand{\best}[1]{\textbf{#1}}
\newcommand{\second}[1]{\underline{#1}}

\newcolumntype{L}[1]{>{\raggedright\arraybackslash}p{#1}}

\newcommand{\ghrepo}[3]{\href{#1}{\footnotesize\path{https://github.com/#2/#3}}}

\journal{Medical Image Analysis}

\begin{document}

\verso{Y. Liu \textit{et~al.}}

\begin{frontmatter}

\title{How Far Has AI Come in Liver Fibrosis Staging? A Real-World Multi-Center Dataset and Benchmark \tnoteref{funding}}

\author[1]{Yuanye \snm{Liu}}
\author[2]{Nannan \snm{Shi}}
\author[3]{Zhejia \snm{Zhang}}
\author[4,5]{Hanxiao \snm{Zhang}}
\author[6]{Boya \snm{Wang}}
\author[7]{Derong \snm{Yu}}
\author[8]{Nao \snm{Wang}}
\author[9]{Yuxin \snm{Jin}}
\author[10]{Yang \snm{Zhou}}
\author[11]{Kunhao \snm{Yuan}}
\author[12]{Siqi \snm{Wang}}
\author[13]{Lida \snm{Yang}}

\author[13]{Xu \snm{Qiao}}
\author[12]{Wentao \snm{Liu}}
\author[9]{Xuelei \snm{He}}
\author[8]{Xin \snm{Hong}}
\author[7]{Guoyan \snm{Zheng}}
\author[6]{Xin \snm{Chen}}
\author[4,5]{Guang-Zhong \snm{Yang}}
\author[14]{Le \snm{Zhang}}
\author[15]{Lei \snm{Li}}
\author[2]{Yuxin \snm{Shi}}
\author[1]{Xiahai \snm{Zhuang}\corref{*}}
\ead{zxh@fudan.edu.cn}
\cortext[*]{Corresponding author}

\address[1]{School of Data Science, Fudan University, Shanghai, China}
\address[2]{Department of Radiology, Shanghai Public Health Clinical Center, Fudan University, Shanghai, China}
\address[3]{Department of Electrical and Computer Engineering, Northwestern University, Evanston, USA}
\address[4]{Shanghai Key Laboratory of Flexible Medical Robotics, Tongren Hospital, Institute of Medical Robotics, Shanghai Jiao Tong University, Shanghai, China}
\address[5]{School of Biomedical Engineering, Shanghai Jiao Tong University, Shanghai, China}
\address[6]{School of Computer Science, University of Nottingham, Nottingham, UK}
\address[7]{Institute of Medical Robotics, School of Biomedical Engineering, Shanghai Jiao Tong University, Shanghai, China}
\address[8]{College of Computer Science and Technology, Huaqiao University, Xiamen, China}
\address[9]{School of Electronic Information (School of Artificial Intelligence), Northwest University, Xi'an, China}
\address[10]{Department of Mechanical Engineering, University College London, London, UK}
\address[11]{Institute of Neuroscience and Cardiovascular Research, University of Edinburgh, Edinburgh, UK}
\address[12]{CAS Center for Excellence in Nanoscience, National Center for Nanoscience and Technology, Beijing, China}
\address[13]{School of Control Science and Engineering, Shandong University, Jinan, China}
\address[14]{School of Engineering, College of Engineering and Physical Sciences, University of Birmingham, Birmingham, UK}
\address[15]{Department of Biomedical Engineering, National University of Singapore, Singapore, Singapore}

\received{4 May 2024}
\finalform{**}
\accepted{**}
\availableonline{**}
\communicated{**}

\begin{abstract}
Despite years of methodological progress, how far AI has come in liver fibrosis staging has never been systematically evaluated under the heterogeneous, multi-center conditions that define clinical practice.
To address this gap, we introduce LiFS (Liver Fibrosis Staging), 
a multi-center dataset and benchmark
derived from the MICCAI 2025 CARE-Liver challenge, comprising 610 patients across multiple centers and scanners with multi-sequence MRI (T1W, T2W, DWI, and four-phase gadoxetic acid-enhanced imaging).
To the best of our knowledge, LiFS is the first benchmark providing complete gadoxetic acid-enhanced sequences with histopathology-confirmed annotations from diverse real-world scanners.
Through systematic evaluation of nine independently developed LiFS methods under a common CARE 2025 setting, and comparison with in-cohort radiologist performance on the same cases, we assess how far current AI has progressed toward clinically meaningful liver fibrosis staging from three complementary perspectives.
First, against radiologists,
the best AI methods approached the performance of the senior radiologist in selected settings and exceeded the junior radiologist on selected endpoints
while median AI performance generally approached junior-radiologist levels.
Second, from a data perspective, cross-center heterogeneity, label imbalance, and contrast-enhanced sequence variability emerge as the dominant challenges for AI methods. 
In particular, the hepatobiliary phase offers valuable functional information but also introduces cross-center variability, creating a practical dilemma in its use and calling for standardized acquisition and generalization-aware modeling.
Third, from a technical perspective, controlled one-factor-at-a-time experiments quantify the effects of spatial registration, input dimensionality, multi-modal fusion strategy, and backbone architecture, providing practical evidence for model design in liver fibrosis staging.
Overall, LiFS provides a rigorous real-world benchmark for positioning the current state of AI in liver fibrosis staging and for enabling future research on the key challenges that limit clinically reliable deployment.
\end{abstract}


\end{frontmatter}

\begin{figure*}[t!]
    \centering
    \includegraphics[width=\textwidth]{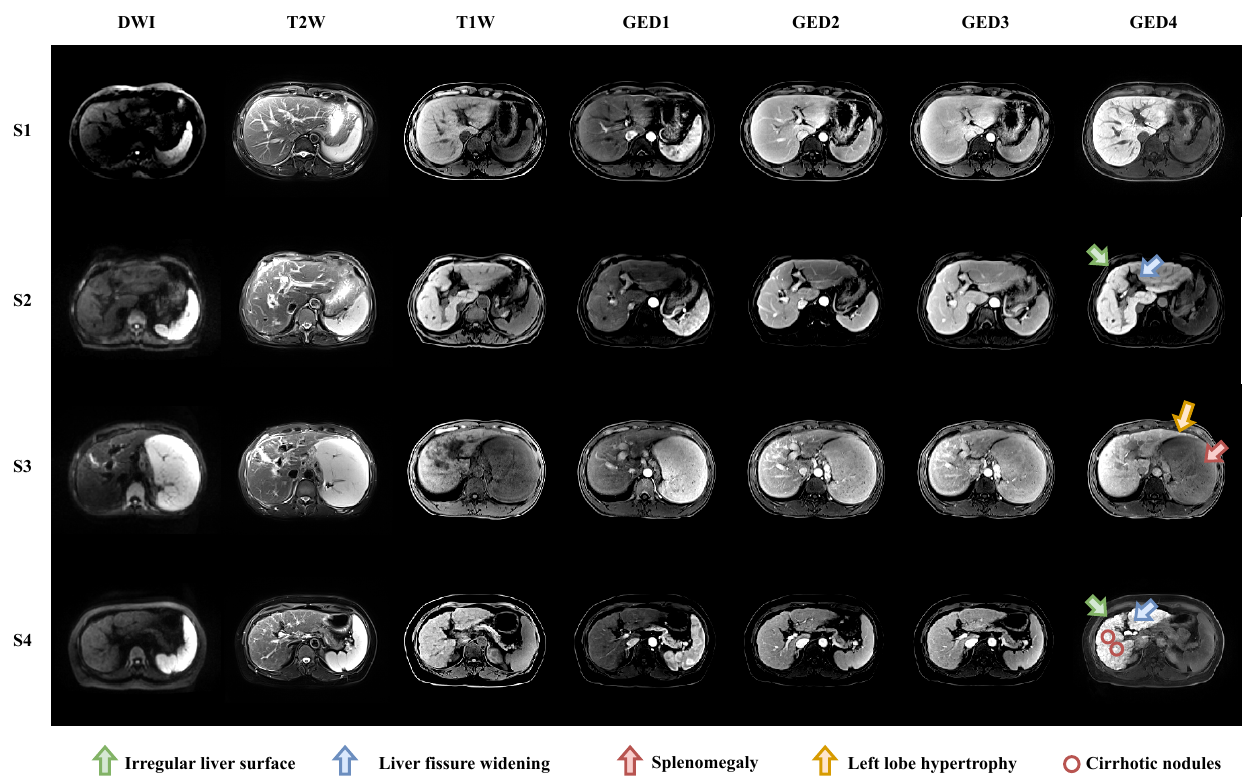}
    \caption{Sample images from the LiFS dataset demonstrating the four fibrosis stages (S1-S4) across seven MRI sequences. Each row represents one fibrosis stage, and each column shows a different MRI sequence: Diffusion-Weighted Imaging (DWI), T2-weighted (T2W), T1-weighted (T1W), and four dynamic phases of gadoxetic acid-enhanced (GED1--4) imaging (arterial, portal venous, delayed, and hepatobiliary phases, respectively). Representative fibrosis signs in the GED4 (HBP) column are highlighted with arrows and circles to illustrate characteristic imaging features at each stage.}
    \label{fig:intro}
\end{figure*}

\begin{table*}[t!]
\centering
\caption{Comparison of LiFS with existing liver-related MRI datasets. LiFS stands out as the largest multi-center dataset featuring comprehensive GED sequences and histopathology-confirmed fibrosis staging labels. The Fibrosis Staging column indicates the type of label: the three-class grade in CirrMRI600+ is a radiological assessment of cirrhosis severity (mild, moderate, severe), whereas LiFS provides histopathological Scheuer stages. Abbreviations: HBP, hepatobiliary phase; Art., arterial phase; PV, portal venous phase; Del., delayed phase; Radiol., radiological; Histol., histopathological; Seg., segmentation.}
\label{tab:benchmark_comparison}
\resizebox{\textwidth}{!}{
\begin{tabular}{lllllllll}
\toprule
\textbf{Dataset} & \textbf{Patients} & \textbf{Volumes} & \textbf{Multi-center} & \textbf{NonContrast Seq.} & \textbf{Contrast Seq.} & \textbf{Fibrosis Staging} & \textbf{Segmentation} & \textbf{Venue} \\
\midrule
Duke Liver~\citep{J_2023_duke_liver} & 105 & 2,146 & \cmark & T1W, T2W, DWI & Art., PV & \textbf{\xmark} & Liver Seg. & Online \\
CirrMRI600+~\citep{J_2025_cirrhosis600+} & 339 & 628 & \xmark & T1W, T2W & PV & Radiol. (3-class) & Liver Seg. & Online \\
LiverHccSeg~\citep{J_2023_liverhccseg} & 46 & $\sim$184 & \xmark & T1W & Art., PV, Del. & \textbf{\xmark} & Tumor/Liver Seg. & Online \\
CHAOS~\citep{J_2021MIA_chaos} & 40 & 120 & \xmark & T1W, T2W & \xmark & \textbf{\xmark} & Organ Seg. & ISBI 2019 \\
\midrule
\textbf{LiFS (Ours)} & \textbf{610} & \textbf{4,174} & \textbf{\cmark} & \textbf{T1W, T2W, DWI} & \textbf{GED (incl. HBP)} & \textbf{Histol. (S1--S4)} & \textbf{Liver Seg.} & \textbf{MICCAI 2025} \\
\bottomrule
\end{tabular}}
\end{table*}

\section{Introduction}

Artificial intelligence (AI) has demonstrated remarkable potential in medical image analysis, and liver fibrosis staging from MRI is no exception.
Over the past decade, deep learning methods have achieved impressive results on this task, with leading approaches reporting high diagnostic accuracy on curated benchmarks~\citep{J_2017Rad_LiFS_MR_GED}.
Yet most existing methods are developed and tested in controlled settings, rather than in the heterogeneous multi-scanner environments encountered in real clinical practice~\citep{J_2021ER_LiFS_MR_T2,J_2024_LiFS_MRI_HBP}.
A fundamental question thus remains unanswered: how far has AI actually come in staging liver fibrosis clinically?
This paper provides a systematic answer through LiFS, a multi-center dataset and benchmark designed to evaluate the current status of AI.

Liver fibrosis, characterized by excessive extracellular matrix deposition leading to architectural distortion, is a critical stage in the progression of chronic liver disease~\citep{J_2005JCI_liver_fibrosis}, and accurate staging is essential for identifying patients at risk of advancing to cirrhosis or hepatocellular carcinoma (HCC)~\citep{J_2019Hepatology_liver_disease,J_2020Cells_liver_fibrosis}.
The current gold standard, histopathological examination via liver biopsy, is invasive and subject to sampling variability~\citep{J_2001NEJM_liver_biopsy,J_2009Hepatology_liver_biopsy_issue}, motivating the search for non-invasive alternatives, such as serum fibrosis scores and liver stiffness measurement by ultrasound-based or magnetic resonance elastography (MRE)~\citep{J_2015CGH_MRE_meta,J_2021JoH_EASL,J_2025Hepatology_AASLD_imaging}.
Magnetic resonance imaging (MRI) offers detailed soft tissue contrast through multiple complementary sequences (T1W, T2W, DWI)~\citep{J_2023_MRF}, and gadoxetic acid-enhanced (GED) imaging further provides hepatocyte-specific functional information critical for fibrosis characterization~\citep{J_2010Radiology_GED_MRI,J_2016_contrast_enhanced_MRI}, with representative LiFS examples shown in Fig.~\ref{fig:intro}.
Deep learning has shown strong potential for automating staging from such multi-sequence data~\citep{J_2017Rad_LiFS_MR_GED,J_2021ER_LiFS_MR_GED_MV,J_2024_LiFS_MR_fusion,C_2025MICCAI_Bayesmm}, but progress has been constrained by the absence of large-scale, multi-center benchmarks that capture real clinical complexity.

Existing datasets for liver MRI remain limited in scale, sequence coverage, or annotation scope (Table~\ref{tab:benchmark_comparison}): none currently provides the combination of multi-center acquisition, complete gadoxetic acid-enhanced (GED) sequences including the hepatobiliary phase (HBP), and histopathology-confirmed fibrosis staging labels needed to evaluate AI under realistic conditions.
Prior benchmarks have focused primarily on segmentation rather than staging, and have not systematically assessed robustness to cross-center domain shift or multi-modal integration challenges.

To address these gaps, we introduce LiFS, a multi-center benchmark derived from the CARE-Liver challenge held in conjunction with MICCAI 2025\footnote{https://zmic.org.cn/care\_challenges/}, designed to evaluate AI algorithms under authentic clinical conditions.
Through systematic evaluation of nine independently developed
LiFS methods under a common CARE 2025 setting,
this work directly examines a single core question that, to the best of our knowledge, has never been addressed in a unified real-world benchmark:
\textit{how far has current AI progressed toward clinically meaningful performance in liver fibrosis staging?}
We approach this question from three complementary angles:
\begin{itemize}
\setlength{\itemsep}{2pt}
\setlength{\parskip}{0pt}
\setlength{\parsep}{0pt}
    \item[\textbf{(1)}] \textit{AI versus clinical experts.} 
    How does the diagnostic performance of current AI methods compare with that of practicing radiologists on the same real-world MRI cases, under both matched acquisition conditions and cross-center distribution shifts?
    \item[\textbf{(2)}] \textit{Data-perspective challenges.} What data-side factors, including cross-center heterogeneity, label imbalance, and contrast-enhanced sequence variability, limit AI performance under real clinical conditions?
    \item[\textbf{(3)}] \textit{Technical-perspective challenges.} Which methodological design choices, including input dimensionality, multi-modal fusion strategy, spatial registration, and backbone architecture, appear associated with robustness under realistic acquisition conditions?
\end{itemize}

To support this investigation, our contributions are as follows.
We introduce LiFS, the first multi-center dataset featuring complete gadoxetic acid-enhanced sequences and histopathology-confirmed Scheuer fibrosis grades~\citep{J_1991JHepatology_Scheuer} from diverse real-world scanners, enabling rigorous evaluation of cross-center generalization.
Building on LiFS, we establish a comprehensive benchmark 
focusing on two clinically important binary tasks derived from the fibrosis staging labels. The benchmark examines the core question from the three angles introduced above, anchored by a clinical reference framework that compares AI methods against in-cohort radiologist performance on the same cases.
Finally, through in-depth analysis of the nine evaluated methods, we provide empirically grounded insights spanning all three angles: AI performance relative to radiologists, the data-side factors that limit real-world deployment (cross-center heterogeneity, label imbalance, and contrast-enhanced sequence variability), and the technical design choices associated with robust performance under heterogeneous acquisition conditions.

\begin{figure*}[t]
    \centering
    \includegraphics[width=\textwidth]{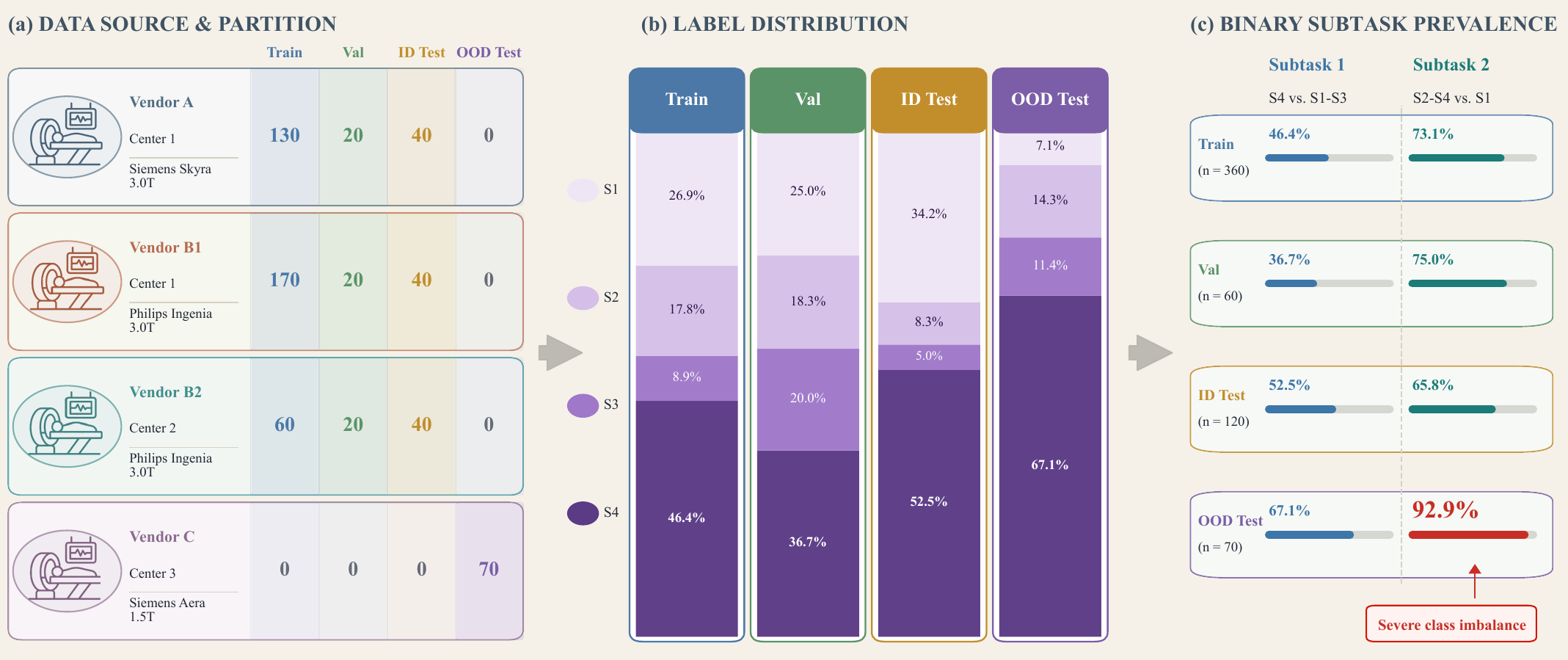}
    \caption{Overview of the LiFS dataset. (a) Data collection and partitioning across multiple centers and vendors. (b) Class distribution of fibrosis stages across training, validation, and test sets. (c) Binary subtask prevalence across train, validation and test sets.}
    \label{fig:vendor_partition}
\end{figure*}

\section{Related Work}
\subsection{Related datasets}

Several public datasets have been established to support research in liver imaging, particularly for tasks like segmentation and disease assessment using MRI.
These resources vary in scale, diversity, and annotation types, but most focus on segmentation rather than explicit fibrosis staging, and they often lack the real-world heterogeneity needed for robust algorithm development.
For instance, the Duke Liver Dataset~\citep{J_2023_duke_liver} comprises 2146 MRI series (including T1W, T2W, DWI, arterial, and portal-venous phases) from 105 patients, many with cirrhotic features. It provides segmentation annotations for the liver but omits fibrosis staging labels and the critical hepatobiliary phase in the contrast-enhanced protocol, limiting its utility for staging tasks. Similarly, CirrMRI600+~\citep{J_2025_cirrhosis600+} includes 628 high-resolution abdominal MRI scans (310 T1W and 318 T2W, with some portal-venous phase data) from 339 patients with liver cirrhosis, offering segmentation masks. However, it is derived from a single center, lacks comprehensive multi-phase GED imaging, and does not include multi-modal sequences like DWI.
Other notable datasets include LiverHccSeg~\citep{J_2023_liverhccseg}, which provides multi-phase MRI (arterial, portal-venous, delayed phases) from 46 patients with HCC, annotated for liver and tumor segmentation but without fibrosis staging. Broader multi-organ MRI datasets like CHAOS~\citep{J_2021MIA_chaos} offer T1W and T2W scans from 40 subjects for abdominal organ segmentation, but they are small-scale, single-center, and not tailored to liver fibrosis or cirrhosis.
Table~\ref{tab:benchmark_comparison} summarizes these comparisons, highlighting how LiFS advances the field with its larger cohort, multi-center acquisition, extensive multi-sequence coverage (including four GED phases), and dedicated fibrosis staging annotations. This enables more realistic benchmarking for non-invasive diagnostic models.

\subsection{Methods for MRI-based liver fibrosis staging}

Deep learning has been applied to liver fibrosis staging across multiple imaging modalities, including ultrasound~\citep{J_2020ER_LiFS_US_1,J_2020ER_LiFS_US_2}, computed tomography~\citep{J_2018Rad_LiFS_CT_1,J_2018ER_LiFS_CT_2}, and histopathology~\citep{C_2021MICCAI_LiFS_WSI_1,C_2024CVPR_LiFS_WSI_2}.
Among these, MRI is particularly attractive owing to its rich tissue contrast and ability to capture functional information without ionizing radiation.
Early MRI-based approaches relied on single sequences such as T2W combined with transfer learning~\citep{J_2021ER_LiFS_MR_T2}, while subsequent works demonstrated the superior diagnostic value of gadoxetic acid-enhanced HBP images~\citep{J_2017Rad_LiFS_MR_GED,J_2021ER_LiFS_MR_GED_MV}.
More recent frameworks integrate T1W, T2W, DWI, and contrast-enhanced phases~\citep{J_2024_LiFS_MR_MulCen,J_2024_LiFS_MR_fusion,J_2024_LiFS_MRI_HBP}, sometimes fusing clinical metadata with imaging features~\citep{J_2025_LiFS_MRI_MulMod}.
Interpretable architectures, such as multi-view evidential learning, have also been explored for this task~\citep{C_2023MICCAI_LiFS_MR_Inter_1,J_2025MIA_LiFS_MR_inter_2}.

However, existing methods have been developed predominantly under controlled, single-center conditions with full-modality availability.
Systematic evaluation of cross-center generalization remains largely absent~\citep{J_2024_LiFS_MR_MulCen}, and interpretability-focused approaches have not been tested under distribution shift.
LiFS aims to fill this gap by providing a unified evaluation platform that exposes the discrepancy between laboratory performance and real-world clinical requirements.

\begin{sidewaystable*}[p]
\centering
\caption{Summary of participants and evaluated methods in the LiFS benchmark. VOI = Volume of Interest; ROI = Region of Interest; CNN = Convolutional Neural Network.}
\label{tab:lifs_participants_methods_summary}
\setlength{\tabcolsep}{3pt}
\renewcommand{\arraystretch}{1.1}
\resizebox{\textheight}{!}{\begin{tabular}{L{2.7cm}L{3.3cm}L{4.5cm}L{5.8cm}L{5.1cm}L{4.2cm}}
\toprule
\textbf{Team} & \textbf{Institute} & \textbf{Data} & \textbf{Method} & \textbf{Training} & \textbf{GitHub Repo} \\
\midrule

SJTU1 \newline {\footnotesize\citep{C_2025CARE_SJTU1}} &
Shanghai Jiao Tong University &
Dim: Cropped Located VOI (3D) \newline
Fusion: Late Fusion (Average) &
Pre-segmentation: Used for VOI cropping \newline
Generalization: Additional pretrain &
Backbone: ResNet-34 (CNN) \newline
Device: NVIDIA 3090 &
\ghrepo{https://github.com/Hanx-Zhang/Liver-Fibrosis-Quantification-and-Analysis}{Hanx-Zhang}{Liver-Fibrosis-Quantification-and-Analysis} \\
\midrule

UoB \newline {\footnotesize\citep{C_2026CARE_UoB}} &
University of Birmingham &
Dim: Whole Volume (3D) \newline
Fusion: Middle fusion (Attention) &
Label imbalance: Stratified sampling \newline
Generalization: Soft voting ensemble \newline
Modality missing: Synthetic compensation &
Backbone: mmFormer (Transformer) \newline
Device: NVIDIA A100 &
\ghrepo{https://github.com/zhangzhejia2002-lgtm/Delta-mmFormer}{zhangzhejia2002-lgtm}{Delta-mmFormer} \\

\midrule

UoN \newline {\footnotesize\citep{C_2026CARE_UoN}} &
University of Nottingham &
Dim: Cropped Located Patch (2D) \newline
Fusion: Early fusion (Patch-based) &
Pre-segmentation: Used for ROI cropping \newline
Label imbalance: Augmentation \newline
Pre-registration: Rigid registration \newline
Modality missing: Zero matrix &
Backbone: ResNet-18 (CNN) \newline
Device: NVIDIA RTX A6000 &
\ghrepo{https://github.com/mileywang3061/Care-Liver}{mileywang3061}{Care-Liver} \\
\midrule

SJTU2 \newline {\footnotesize\citep{C_2026CARE_SJTU2}} &
Shanghai Jiao Tong University &
Dim: Overlapping Volume (3D) \newline
Fusion: Middle fusion (Attention) &
Label imbalance: Weighted cross-entropy \newline
Modality missing: Random Dropout Train &
Backbone: Anatomix (CNN) \newline
Device: NVIDIA Tesla H100 &
\ghrepo{https://github.com/Kendra9999/MA2/}{Kendra9999}{MA2} \\
\midrule

SDU \newline {\footnotesize\citep{C_2026CARE_SDU}} &
Shandong University &
Dim: Whole Volume (3D) \newline
Fusion: Middle fusion (feature) &
Pre-segmentation: Used for VOI cropping \newline
Modality missing: Zero matrix &
Backbone: ALMSS (CNN) \newline
Device: NVIDIA A100 &
\ghrepo{https://github.com/LidaY412/34-LiFS}{LidaY412}{34-LiFS} \\
\midrule

NCTST \newline {\footnotesize\citep{C_2026CARE_NCTST}} &
National Center for Nanoscience and Technology &
Dim: Whole Volume (3D) \newline
Fusion: Middle fusion (feature)  &
Pre-registration: \cmark \newline
Modality missing: Zero matrix &
Backbone: UniFormer (Transformer) \newline
Device: NVIDIA RTX 8000 &
\ghrepo{https://github.com/WWangSiqi/Multibranch-attention-Network}{WWangSiqi}{Multibranch-attention-Network} \\
\midrule

HQU \newline {\footnotesize\citep{C_2026CARE_HQU}} &
Huaqiao University &
Dim: Compressed Volume (2D, described as 2.5D in paper) \newline
Fusion: Middle fusion (feature) &
Label imbalance: Focal Loss \newline
Modality missing: Zero matrix &
Backbone: Swin Transformer \newline
Device: NVIDIA RTX 4090 &
\ghrepo{https://github.com/hxpotato/CARE2025}{hxpotato}{CARE2025} \\
\midrule

NWU-CN \newline {\footnotesize\citep{C_2026CARE_NWUCN}} &
Northwest University &
Dim: Cropped Located Patch (2D) \newline
Fusion: Late fusion (attention gating) &
Pre-segmentation: Used for ROI cropping \newline
Generalization: Uncertainty-guided curriculum learning &
Backbone: ResNet-18 (CNN) \newline
Device: NVIDIA RTX 4090
 &
\ghrepo{https://github.com/pazjin/FibUCL}{pazjin}{FibUCL} \\
\midrule

UCL \newline {\footnotesize\citep{C_2026CARE_UCL}} &
University College London &
Dim: Handcrafted Features (32D) \newline
Fusion: Middle fusion &
Pre-segmentation: Used for ROI cropping \newline
Pre-registration: Rigid \newline
Generalization: Additional scanner flag &
Backbone: Random Forest \newline
Device: AMD Ryzen 5 5600 \& NVIDIA A100 &
\ghrepo{https://github.com/YangForever/care2025_liver_biodreamer}{YangForever}{care2025_liver_biodreamer} \\
\bottomrule
\end{tabular}
}
\end{sidewaystable*}

\section{LiFS: A Multi-Modality Liver Fibrosis Staging Dataset}

\subsection{Overview}
We release LiFS, a multi-center, open-access dataset for liver fibrosis staging that reflects real-world clinical complexity.
The dataset includes $610$ patients from multiple centers, scanned with multi-sequence, contrast-enhanced liver MRI.
In total, LiFS contains $4,174$ 3D MRI volumes (in NIfTI format) and $218,203$ slices.
The training and validation subsets, including annotations, are publicly available at {\url{https://zmic.org.cn/care_challenges/}}.
LiFS extends the CARE2024/LiQA dataset by retaining all $440$ original cases and adding $170$ new cases, with unchanged task definitions and annotation protocol.

As summarized in Fig.~\ref{fig:vendor_partition}, data were collected from three clinical centers in China and four different MRI scanners, covering both 1.5T and 3.0T magnetic field strengths.
Vendor A and Vendor B (two centers, three scanners) provided 360, 60 and 120 subjects for training, validation and in-distribution (ID) testing, respectively, while Vendor C (one independent scanner) served exclusively as the out-of-distribution (OOD) test set with 70 patients, used to assess model generalization.
We consider the OOD shift to arise from combined differences in clinical center, scanner, field strength, acquisition protocol, and fibrosis-stage composition.
The cohort was derived from routine clinical practice and included patients with coexisting liver lesions, including HCC.
Across all $610$ patients, age ranged from $16$ to $92$ years. Mean age was similar in the in-distribution and out-of-distribution cohorts, whereas the proportion of male patients was higher in the out-of-distribution cohort.

The MRI acquisition protocol followed the standard multi-phase dynamic contrast-enhanced sequence:
the arterial phase is captured approximately 25 seconds after contrast injection, followed by the portal venous phase (after 1 minute), delayed phase (after an additional 90 seconds), and finally the HBP acquired 20 minutes post-injection.
HBP has been shown to provide functional liver signals most informative for fibrosis characterization.

Ground-truth was determined by histopathological examination from biopsy or surgical resection performed within three months of the MRI scan, ensuring clinical consistency and reliability. 
Fibrosis stages were defined according to the Scheuer grading system~\citep{J_1991JHepatology_Scheuer}: S1 (fibrous portal expansion), S2 (periportal fibrosis), S3 (fibrosis with architectural distortion), and S4 (cirrhosis). 

\subsection{Real-world features}
LiFS was designed to replicate the real-world challenges encountered in clinical MRI data analysis, providing a comprehensive platform for developing robust and generalizable algorithms.
Key characteristics include:
(1) Spatial Misalignment: The multi-sequence MRI volumes (T1W, T2W, DWI, GED1--4) are not spatially registered, reflecting routine acquisition variability in clinical workflows.
(2) Missing Modalities: Certain subjects lack one or more sequences due to acquisition failure, motion artifacts, or protocol variations, making LiFS suitable for evaluating modality-robust learning and modality completion methods.
(3) Cross-center Heterogeneity and Generalization: Data were collected from three hospitals and four scanners using different vendors and field strengths, introducing domain shifts that test the cross-site generalization ability of learning algorithms.

It is worth noting that the HBP, acquired approximately 20 minutes post-injection, is particularly susceptible to inter-center variability.
Differences in injection protocols, patient liver function, and scanner characteristics may substantially alter HBP signal intensity across sites, introducing domain shifts that are qualitatively distinct from those observed in routine anatomical sequences. 
This property makes LiFS uniquely suited to evaluate whether AI methods can robustly exploit functional hepatobiliary information across heterogeneous acquisition environments.

\subsection{Research opportunities}

To the best of our knowledge, LiFS is the largest annotated multi-sequence MRI dataset for liver fibrosis staging to date.
Its scale, diversity, and real-world imperfections open a wide range of research directions, including but not limited to:
(1) Multi-modality fusion and representation learning;
(2) Domain generalization and adaptation across centers and vendors;
(3) Missing-modality reconstruction and modality-agnostic inference;
(4) Robustness and uncertainty estimation for clinical reliability.
We welcome the community to explore and build upon LiFS to advance the development of reliable, interpretable, and clinically applicable AI for liver disease assessment.

\section{CARE-Liver Challenge}

\subsection{Challenge set-up}
The CARE Challenge was held in conjunction with MICCAI 2025, aiming to address the comprehensive challenges inherent to real-world clinical deployment.
Among all tasks, the CARE-Liver track focuses on liver MRI, comprising two tasks: LiFS and Liver Segmentation (LiSeg).
For LiFS, we established two experimental tracks: (1) Contrast-Enhanced, utilizing all available sequences (T1W, T2W, DWI, and GED1--4); and (2) Non-Contrast, restricted to T1W, T2W, and DWI sequences to assess performance without exogenous agents.
Two clinically important subtasks, Cirrhosis Detection (S4 vs. S1-3) and Significant Fibrosis Detection (S1 vs. S2-4)~\citep{J_2018Rad_LiFS_CT_1,J_2021JoH_EASL}, were evaluated.
In both subtasks the positive label is a single Scheuer stage: S4 for Subtask~1 and S1 for Subtask~2.
These binary endpoints constitute the primary evaluation tasks of the current LiFS benchmark.

The CARE-Liver track attracted $96$ registered teams and received $220$ evaluation submissions.
Among the $13$ papers accepted in the CARE 2025 workshop proceedings across LiFS and LiSeg~\citep{B_2025CARE}, eight presented LiFS methods included in this benchmark.
We additionally included SJTU1, originally published in the CARE 2024 proceedings~\citep{C_2025CARE_SJTU1}, and re-evaluated it under the same CARE 2025 setting.
Table~\ref{tab:lifs_participants_methods_summary} summarizes the participating teams, their affiliations, the main design characteristics of their methods, and the availability of released code repositories.

\subsection{Evaluation}\label{subsec:eval}
This benchmark is reported in accordance with the CLAIM 2024 update checklist~\citep{J_2024RadiologyAI_CLAIM}.
Model-specific development details are provided in the corresponding method papers listed in Table~\ref{tab:lifs_participants_methods_summary}.

We evaluated all methods on both in-distribution (ID) and OOD test sets. The ID test data were drawn from the same distribution as the training set, whereas the OOD test data came from a completely unseen scanner and clinical center (Vendor C), enabling assessment of cross-domain generalization.

The challenge's official evaluation metrics were classification accuracy (ACC) and area under the receiver operating characteristic (ROC) curve (AUC).
ACC reflects overall classification correctness, while AUC provides a threshold-independent measure of discriminative ability, which is particularly important under label imbalance.
Evaluation scripts and baseline implementations are publicly available through the LiFS Toolkit repository (\url{https://github.com/HenryLau7/LiFS-toolkit}), facilitating reproducible evaluation and future method development.

In addition to the nine evaluated methods, we implemented two baselines, ResNet3D~\citep{C_2016CVPR_resnet} and ViT3D~\citep{C_2021ICLR_vit}, to provide standardized reference models under a controlled training protocol.
Both baselines were trained by the organizers using the same official training and validation partitions as the challenge participants and were evaluated on the same ID and OOD test sets. Their detailed configurations are provided in Sec.~\ref{subsec:design_choices}.

\paragraph{Radiologist annotation}\label{para:expert_annotation}
To establish an in-cohort clinical reference, two board-certified radiologists with $3$ and $8$ years of abdominal imaging experience (denoted Expert-3Y and Expert-8Y, respectively) served as in-cohort clinical readers and independently evaluated the same LiFS benchmark cases using the 7-point qualitative MRI scoring scheme of~\citet{J_2015_7scale}, in which an overall qualitative grading for the presence of fibrosis is assigned on a $1$--$7$ scale.
Both radiologists were blinded to the histopathological ground truth, patient clinical information, the other radiologist's scores, and the train/validation/test partition (including the ID/OOD designation).
Each radiologist completed two reading sessions with case order independently randomized by reader and session.
All cases were first assessed using only the non-contrast sequences (T1W, T2W, DWI).
After an interval of approximately $2$ hours, the same cases were assessed with the additional contrast-enhanced sequences.
Both readers showed an overall tendency toward higher scores after contrast-enhanced sequences were introduced, indicating that the additional sequences contributed to their fibrosis assessments.
The two readings were retained strictly as separate references (Expert-3Y and Expert-8Y) and no averaging or consensus reconciliation was performed.
Because the radiologist ratings were ordinal, AUC was used as the primary metric for comparison with AI methods, retaining the ranking information of the 7-point scores.

\paragraph{Statistical analysis}
Two-sided $95\%$ DeLong confidence intervals were computed for all AUCs.
Within each evaluation cell, the highest-AUC submission (Best AI) and median-ranked submission (Median AI) were compared with each radiologist using paired DeLong tests on the case-aligned cohort, with Benjamini--Hochberg correction applied across the resulting $32$ comparisons.
Inter-reader agreement was assessed using quadratic-weighted Cohen's $\kappa$ on the 7-point scores and unweighted $\kappa$ on the binarised Subtask~1 endpoint, with $95\%$ confidence intervals obtained from $2{,}000$ case-level bootstrap resamples.

To separate the contribution of fibrosis-stage composition from the composite OOD shift, we used stage-standardised AUC. 
For each method, the ID negatives were reweighted to match the OOD distribution of negative stages, while the positive class was left unchanged. 
The resulting $\mathrm{AUC}_{\mathrm{ID,std}}$ therefore estimates ID performance under the OOD stage composition while retaining the ID cases and acquisition environment. 
We defined $\Delta_{\mathrm{total}}=\mathrm{AUC}_{\mathrm{ID}}-\mathrm{AUC}_{\mathrm{OOD}}$, $\Delta_{\mathrm{case\mbox{-}mix}}=\mathrm{AUC}_{\mathrm{ID}}-\mathrm{AUC}_{\mathrm{ID,std}}$, and $\Delta_{\mathrm{residual}}=\mathrm{AUC}_{\mathrm{ID,std}}-\mathrm{AUC}_{\mathrm{OOD}}$. Thus, $\Delta_{\mathrm{case\mbox{-}mix}}$ captures the component associated with the measured difference in stage composition, whereas $\Delta_{\mathrm{residual}}$ is the remaining non-stage component after this standardisation, combining center, scanner, field strength, protocol, and any other unmeasured cohort differences. 
Confidence intervals were obtained from $2{,}000$ stage-by-class-stratified bootstrap resamples with fixed standardisation weights. Sensitivity analyses used an exactly stage-matched ID subset and subtracted each radiologist’s ID–OOD difference from each method’s difference. We additionally calculated stage-standardised AUC separately for each scanner cohort.

\begin{table*}[!t]
\centering
\caption{
Discrimination performance on the LiFS benchmark.
Each method and reader cell reports AUC ($\times100$) with its two-sided $95\%$ DeLong confidence interval; the Best AI and Median AI summary rows report AUC only.
Subtask~1 and Subtask~2 denote S4 vs.\ S1--3 and S1 vs.\ S2--4, respectively; ID and OOD denote the in-distribution and out-of-distribution test partitions.
Positive and negative case counts are given in the column headers.
Best and second-best results among the nine evaluated methods are shown in \textbf{bold} and \underline{underlined}, respectively.
Best AI and Median AI are determined separately for each evaluation cell, with subscripts identifying the corresponding team.
Superscript arrows indicate significant paired DeLong comparisons with the specified reader: $^{\ast}$ denotes significance after Benjamini--Hochberg correction ($q<0.05$), whereas $^{\dagger}$ denotes significance before correction only ($p<0.05$, $q\geq0.05$).
Results shaded in \colorbox{gray!10}{grey} were evaluated after the challenge closed.
Baseline backbones were trained under a common controlled configuration and are reported as five-seed probability ensembles.
}
\label{tab:results_merged}
\setlength{\tabcolsep}{4pt}\renewcommand{\arraystretch}{1.22}
\resizebox{\textwidth}{!}{\small
\renewcommand{\cellset}{\renewcommand{\arraystretch}{0.8}}
\newcolumntype{N}{>{\centering\arraybackslash}m{46pt}}
\begin{tabular}{@{}l@{\hspace{10pt}}NNNN!{\color{black!25}\vrule width .4pt}NNNN@{}}
\toprule
& \multicolumn{4}{c}{\cellcolor{cebg}\textbf{Contrast-Enhanced}}
& \multicolumn{4}{c}{\cellcolor{ncbg}\textbf{Non-Contrast}} \\
\cmidrule(lr){2-5}\cmidrule(lr){6-9}
& \multicolumn{2}{c}{\textbf{ID}} & \multicolumn{2}{c}{\textbf{OOD}}
& \multicolumn{2}{c}{\textbf{ID}} & \multicolumn{2}{c}{\textbf{OOD}} \\
\cmidrule(lr){2-3}\cmidrule(lr){4-5}\cmidrule(lr){6-7}\cmidrule(lr){8-9}
\makecell[l]{\textbf{Method}} & \makecell{Subtask 1\\[1pt]\tabci{$n_{+}{=}63$}\\[-2pt]\tabci{$n_{-}{=}57$}} & \makecell{Subtask 2\\[1pt]\tabci{$n_{+}{=}41$}\\[-2pt]\tabci{$n_{-}{=}79$}} & \makecell{Subtask 1\\[1pt]\tabci{$n_{+}{=}47$}\\[-2pt]\tabci{$n_{-}{=}23$}} & \makecell{Subtask 2\\[1pt]\tabci{$n_{+}{=}5$}\\[-2pt]\tabci{$n_{-}{=}65$}} & \makecell{Subtask 1\\[1pt]\tabci{$n_{+}{=}63$}\\[-2pt]\tabci{$n_{-}{=}57$}} & \makecell{Subtask 2\\[1pt]\tabci{$n_{+}{=}41$}\\[-2pt]\tabci{$n_{-}{=}79$}} & \makecell{Subtask 1\\[1pt]\tabci{$n_{+}{=}47$}\\[-2pt]\tabci{$n_{-}{=}23$}} & \makecell{Subtask 2\\[1pt]\tabci{$n_{+}{=}5$}\\[-2pt]\tabci{$n_{-}{=}65$}} \\
\midrule
\makecell[l]{ResNet3D} & \cellcolor{gray!10}\makecell{77.81\\[1pt]\tabci{(69.2--86.4)}} & \cellcolor{gray!10}\makecell{78.45\\[1pt]\tabci{(69.7--87.2)}} & \cellcolor{gray!10}\makecell{58.19\\[1pt]\tabci{(43.4--73.0)}} & \cellcolor{gray!10}\makecell{54.46\\[1pt]\tabci{(17.7--91.2)}} & \cellcolor{gray!10}\makecell{80.06\\[1pt]\tabci{(71.7--88.4)}} & \cellcolor{gray!10}\makecell{72.95\\[1pt]\tabci{(63.4--82.5)}} & \cellcolor{gray!10}\makecell{61.05\\[1pt]\tabci{(47.1--75.0)}} & \cellcolor{gray!10}\makecell{52.92\\[1pt]\tabci{(20.6--85.2)}} \\
\makecell[l]{ViT3D} & \cellcolor{gray!10}\makecell{74.58\\[1pt]\tabci{(65.8--83.3)}} & \cellcolor{gray!10}\makecell{78.14\\[1pt]\tabci{(69.2--87.0)}} & \cellcolor{gray!10}\makecell{68.36\\[1pt]\tabci{(55.9--80.8)}} & \cellcolor{gray!10}\makecell{65.54\\[1pt]\tabci{(38.8--92.3)}} & \cellcolor{gray!10}\makecell{75.72\\[1pt]\tabci{(66.9--84.5)}} & \cellcolor{gray!10}\makecell{66.87\\[1pt]\tabci{(56.2--77.5)}} & \cellcolor{gray!10}\makecell{57.08\\[1pt]\tabci{(41.8--72.4)}} & \cellcolor{gray!10}\makecell{48.62\\[1pt]\tabci{(16.1--81.2)}} \\
\midrule[\lightrulewidth]
\makecell[l]{UCL} & \cellcolor{gray!10}\makecell{81.83\\[1pt]\tabci{(74.0--89.7)}} & \cellcolor{gray!10}\makecell{81.83\\[1pt]\tabci{(72.8--90.9)}} & \cellcolor{gray!10}\makecell{\best{77.98}\\[1pt]\tabci{(66.3--89.7)}} & \cellcolor{gray!10}\makecell{50.62\\[1pt]\tabci{(37.6--63.6)}} & \cellcolor{gray!10}\makecell{81.54\\[1pt]\tabci{(73.7--89.4)}} & \cellcolor{gray!10}\makecell{\best{81.75}\\[1pt]\tabci{(73.8--89.7)}} & \cellcolor{gray!10}\makecell{56.89\\[1pt]\tabci{(41.5--72.3)}} & \cellcolor{gray!10}\makecell{58.62\\[1pt]\tabci{(26.7--90.5)}} \\
\makecell[l]{SJTU2} & \makecell{75.10\\[1pt]\tabci{(66.2--84.0)}} & \makecell{78.45\\[1pt]\tabci{(69.7--87.2)}} & \makecell{46.81\\[1pt]\tabci{(32.1--61.5)}} & \makecell{52.62\\[1pt]\tabci{(23.1--82.1)}} & \makecell{78.61\\[1pt]\tabci{(70.4--86.8)}} & \makecell{73.70\\[1pt]\tabci{(63.7--83.7)}} & \makecell{49.31\\[1pt]\tabci{(33.6--65.0)}} & \makecell{31.08\\[1pt]\tabci{(9.2--52.9)}} \\
\makecell[l]{NWU-CN} & \makecell{78.03\\[1pt]\tabci{(69.8--86.2)}} & \makecell{78.20\\[1pt]\tabci{(69.6--86.8)}} & \makecell{64.11\\[1pt]\tabci{(49.8--78.4)}} & \makecell{52.62\\[1pt]\tabci{(20.2--85.0)}} & \makecell{79.78\\[1pt]\tabci{(71.6--88.0)}} & \makecell{67.27\\[1pt]\tabci{(56.6--78.0)}} & \makecell{46.25\\[1pt]\tabci{(31.2--61.3)}} & \makecell{40.31\\[1pt]\tabci{(16.6--64.0)}} \\
\makecell[l]{UoN} & \makecell{\second{83.92}\\[1pt]\tabci{(76.8--91.0)}} & \makecell{\second{83.42}\\[1pt]\tabci{(75.7--91.1)}} & \makecell{54.12\\[1pt]\tabci{(39.4--68.8)}} & \makecell{\second{75.38}\\[1pt]\tabci{(50.4--100.0)}} & \makecell{77.22\\[1pt]\tabci{(68.4--86.1)}} & \makecell{74.28\\[1pt]\tabci{(65.6--83.0)}} & \makecell{\second{69.47}\\[1pt]\tabci{(56.0--82.9)}} & \makecell{\best{86.31}\\[1pt]\tabci{(71.4--100.0)}} \\
\makecell[l]{HQU} & \cellcolor{gray!10}\makecell{73.77\\[1pt]\tabci{(64.5--83.0)}} & \cellcolor{gray!10}\makecell{79.47\\[1pt]\tabci{(70.8--88.2)}} & \cellcolor{gray!10}\makecell{50.23\\[1pt]\tabci{(34.7--65.7)}} & \cellcolor{gray!10}\makecell{52.31\\[1pt]\tabci{(22.1--82.5)}} & \makecell{77.22\\[1pt]\tabci{(68.5--86.0)}} & \makecell{78.51\\[1pt]\tabci{(70.4--86.7)}} & \makecell{\best{71.51}\\[1pt]\tabci{(58.7--84.3)}} & \makecell{33.23\\[1pt]\tabci{(0.0--66.9)}} \\
\makecell[l]{UoB} & \cellcolor{gray!10}\makecell{71.85\\[1pt]\tabci{(62.6--81.1)}} & \cellcolor{gray!10}\makecell{78.14\\[1pt]\tabci{(69.2--87.1)}} & \cellcolor{gray!10}\makecell{\second{73.54}\\[1pt]\tabci{(61.1--86.0)}} & \cellcolor{gray!10}\makecell{\best{79.69}\\[1pt]\tabci{(62.8--96.6)}} & \makecell{71.73\\[1pt]\tabci{(62.3--81.1)}} & \makecell{68.48\\[1pt]\tabci{(58.4--78.6)}} & \makecell{68.83\\[1pt]\tabci{(56.3--81.4)}} & \makecell{\second{71.38}\\[1pt]\tabci{(51.7--91.1)}} \\
\makecell[l]{SDU} & \cellcolor{gray!10}\makecell{80.62\\[1pt]\tabci{(72.7--88.6)}} & \cellcolor{gray!10}\makecell{76.57\\[1pt]\tabci{(67.1--86.1)}} & \cellcolor{gray!10}\makecell{61.05\\[1pt]\tabci{(47.4--74.7)}} & \cellcolor{gray!10}\makecell{54.46\\[1pt]\tabci{(32.5--76.5)}} & \makecell{\best{83.76}\\[1pt]\tabci{(76.5--91.0)}} & \makecell{68.94\\[1pt]\tabci{(56.9--80.9)}} & \makecell{68.73\\[1pt]\tabci{(56.1--81.4)}} & \makecell{70.15\\[1pt]\tabci{(53.3--87.0)}} \\
\makecell[l]{NCTST} & \makecell{78.31\\[1pt]\tabci{(69.6--87.1)}} & \makecell{\best{84.50}\\[1pt]\tabci{(76.3--92.7)}} & \makecell{66.05\\[1pt]\tabci{(52.4--79.7)}} & \makecell{37.23\\[1pt]\tabci{(5.1--69.3)}} & \cellcolor{gray!10}\makecell{79.94\\[1pt]\tabci{(71.2--88.7)}} & \cellcolor{gray!10}\makecell{\second{78.96}\\[1pt]\tabci{(69.8--88.1)}} & \cellcolor{gray!10}\makecell{58.74\\[1pt]\tabci{(44.3--73.1)}} & \cellcolor{gray!10}\makecell{60.00\\[1pt]\tabci{(27.1--92.9)}} \\
\makecell[l]{SJTU1} & \cellcolor{gray!10}\makecell{\best{86.16}\\[1pt]\tabci{(79.7--92.6)}} & \cellcolor{gray!10}\makecell{77.00\\[1pt]\tabci{(68.6--85.4)}} & \cellcolor{gray!10}\makecell{49.40\\[1pt]\tabci{(34.7--64.1)}} & \cellcolor{gray!10}\makecell{68.62\\[1pt]\tabci{(52.5--84.7)}} & \cellcolor{gray!10}\makecell{\second{82.12}\\[1pt]\tabci{(74.7--89.5)}} & \cellcolor{gray!10}\makecell{76.32\\[1pt]\tabci{(67.7--84.9)}} & \cellcolor{gray!10}\makecell{45.42\\[1pt]\tabci{(30.6--60.3)}} & \cellcolor{gray!10}\makecell{64.92\\[1pt]\tabci{(45.3--84.6)}} \\
\midrule[\lightrulewidth]
\makecell[l]{Expert-3Y} & \cellcolor{gray!10}\makecell{73.63\\[1pt]\tabci{(65.4--81.9)}} & \cellcolor{gray!10}\makecell{76.51\\[1pt]\tabci{(68.6--84.4)}} & \cellcolor{gray!10}\makecell{63.88\\[1pt]\tabci{(51.0--76.8)}} & \cellcolor{gray!10}\makecell{52.46\\[1pt]\tabci{(35.1--69.8)}} & \cellcolor{gray!10}\makecell{69.53\\[1pt]\tabci{(60.8--78.2)}} & \cellcolor{gray!10}\makecell{69.76\\[1pt]\tabci{(61.0--78.5)}} & \cellcolor{gray!10}\makecell{54.81\\[1pt]\tabci{(41.8--67.8)}} & \cellcolor{gray!10}\makecell{64.31\\[1pt]\tabci{(44.2--84.4)}} \\
\makecell[l]{Expert-8Y} & \cellcolor{gray!10}\makecell{81.83\\[1pt]\tabci{(74.3--89.4)}} & \cellcolor{gray!10}\makecell{80.07\\[1pt]\tabci{(71.8--88.4)}} & \cellcolor{gray!10}\makecell{66.65\\[1pt]\tabci{(54.8--78.5)}} & \cellcolor{gray!10}\makecell{71.38\\[1pt]\tabci{(53.0--89.8)}} & \cellcolor{gray!10}\makecell{77.29\\[1pt]\tabci{(68.8--85.7)}} & \cellcolor{gray!10}\makecell{74.96\\[1pt]\tabci{(66.1--83.8)}} & \cellcolor{gray!10}\makecell{77.57\\[1pt]\tabci{(66.9--88.3)}} & \cellcolor{gray!10}\makecell{78.92\\[1pt]\tabci{(67.7--90.2)}} \\
\midrule[\lightrulewidth]
\makecell[l]{Best AI} & $86.16_{\mathrm{SJTU1}}^{\uparrow\mathrm{3Y},\ast}$ & $84.50_{\mathrm{NCTST}}$ & $77.98_{\mathrm{UCL}}$ & $79.69_{\mathrm{UoB}}^{\uparrow\mathrm{3Y},\dagger}$ & $83.76_{\mathrm{SDU}}^{\uparrow\mathrm{3Y},\ast}$ & $81.75_{\mathrm{UCL}}^{\uparrow\mathrm{3Y},\dagger}$ & $71.51_{\mathrm{HQU}}$ & $86.31_{\mathrm{UoN}}$ \\
\makecell[l]{Median AI} & $78.31_{\mathrm{NCTST}}$ & $78.45_{\mathrm{SJTU2}}$ & $61.05_{\mathrm{SDU}}$ & $52.62_{\mathrm{SJTU2}}$ & $79.78_{\mathrm{NWU-CN}}^{\uparrow\mathrm{3Y},\dagger}$ & $74.28_{\mathrm{UoN}}$ & $58.74_{\mathrm{NCTST}}^{\downarrow\mathrm{8Y},\dagger}$ & $60.00_{\mathrm{NCTST}}$ \\
\bottomrule
\end{tabular}
}
\end{table*}

\subsection{Methodological overview}
As summarized in Table \ref{tab:lifs_participants_methods_summary}, the nine evaluated methods exhibit a diverse array of strategies tailored to the challenges of liver fibrosis staging. We categorize these contributions along three primary dimensions: input representation and backbone architecture, multi-modal fusion strategy, and mechanisms for handling real-world heterogeneity.

\textbf{Input Representation and Backbone Architecture.}
Volumetric 3D approaches were the predominant choice, adopted by five teams (SJTU1, UoB, SJTU2, SDU, NCTST) to exploit global contextual information from the liver MRI.
In contrast, 2D patch-based networks (UoN, NWU-CN) focused on capturing fine-grained local textural details, while Team HQU employed a 2.5D compression strategy to balance spatial context with computational efficiency.
Notably, one team (UCL) utilized handcrafted radiomics features (32D) rather than deep feature extraction, serving as a classical baseline.
Regarding backbone architectures, a transition towards Transformer-based models is observed, with three teams deploying advanced architectures such as mmFormer~\citep{C_2022MICCAI_mmformer}, UniFormer~\citep{J_2023PAMI_WSQBackbone}, and Swin Transformer~\citep{C_2021ICCV_swintrans} to capture long-range dependencies, while others relied on established CNN baselines like ResNet~\citep{C_2016CVPR_resnet}, ALMSS~\citep{J_2024_LidaYangBackbone} and Anatomix~\citep{C_2025ICLR_anatomix}.

\textbf{Multi-modal Fusion Strategy.}
Effective integration of multi-sequence MRI is critical for this task. Middle-level fusion emerges as the dominant strategy (adopted by 6 teams), typically utilizing attention mechanisms or cross-modal modules to interactively learn sequence-level features.
Early fusion was employed by UoN, concatenating modality channels or patches at the input stage to learn joint representations from raw data.
Conversely, late fusion strategies (SJTU1, NWU-CN) aggregated modality-wise predictions via averaging or gating mechanisms, prioritizing independent feature extraction before final decision-making.

\textbf{Handling Real-world Challenges.}
Participants introduced specific modules to address the intrinsic data imperfections of the LiFS benchmark.
\textit{(1) Spatial Misalignment:} To mitigate the lack of pixel-wise alignment across sequences, three teams (UoN, NCTST, UCL) explicitly incorporated rigid registration as a pre-processing step.
\textit{(2) Missing Modalities:} To ensure robustness against incomplete scans, several methods incorporated modality-agnostic training strategies. These included zero-padding (UoN, SDU, HQU, NCTST), random modality dropout during training (SJTU2), and synthetic compensation (UoB).
\textit{(3) Generalization \& Imbalance:} Advanced learning paradigms were explored to enhance OOD generalization, such as uncertainty-guided curriculum learning (NWU-CN) and soft voting ensembles (UoB). Furthermore, label imbalance was addressed through stratified sampling (UoB), weighted cross-entropy (SJTU2), and focal loss (HQU).

\begin{figure*}[th!]
    \centering
    \includegraphics[width=\textwidth]{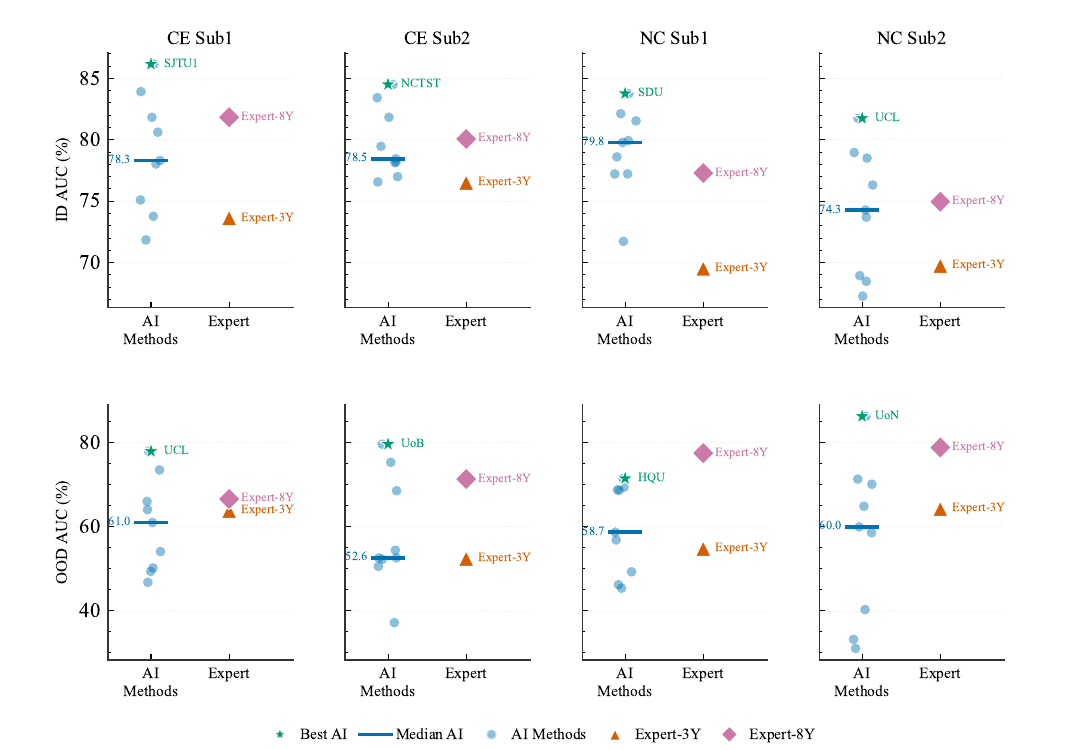}
    \caption{AUC comparison between radiologists and AI teams across ID/OOD and contrast-enhanced/non-contrast settings. Each blue circle represents an individual AI team, the horizontal bar indicates the median AI performance, and the star highlights the best AI performance. Red triangles and gold diamonds represent Expert-3Y and Expert-8Y, respectively. The plot illustrates both the best-case AI performance and the dispersion of AI results relative to in-cohort radiologist references. Expert-3Y and Expert-8Y denote radiologists with 3 and 8 years of experience, respectively.}
    \label{fig:expert_ai_auc}
\end{figure*}

\section{Results}

\subsection{AI methods vs. clinical reference: how far has AI progressed toward radiologist-level performance?}\label{subsec:ai_vs_expert}

To evaluate current AI progress relative to clinical references on the LiFS benchmark, we compared their performance across acquisition conditions (Contrast-Enhanced and Non-Contrast) and ID/OOD evaluation settings, and compared the results with an in-cohort reading by two radiologists with different experience levels (Table~\ref{tab:results_merged}).
We focus here on AUC as the primary evaluation metric.
The leading method varies substantially across different settings, indicating that no single AI strategy dominates the benchmark.
For example, SJTU1 achieved the highest Contrast-Enhanced ID AUC for Subtask 1, reaching $86.16\%$, but its performance dropped markedly on the corresponding OOD test set, with AUC decreasing to $49.40\%$. 
UoN showed the highest Non-Contrast OOD AUC for Subtask 2, $86.31\%$, however, this endpoint rests on only $5$ positive cases against $65$ negatives and the estimate carries a $95\%$ confidence interval of $[71.37, 100.00]$.

\textbf{Under ID evaluation.}
As detailed in Section~\ref{subsec:eval}, the in-cohort reference reading by Expert-3Y and Expert-8Y is reported on the 7-point ordinal scale, with AUC as the primary metric for AI-versus-radiologist comparison on the case-aligned cohort.
As shown in Table~\ref{tab:results_merged} and Fig.~\ref{fig:expert_ai_auc}, the best-performing AI methods achieved strong performance under ID evaluation, with AUCs above the Expert-3Y reference and in the range of the Expert-8Y reference across all four ID settings.
In the contrast-enhanced setting, the best AI achieved AUCs of $86.16$ and $84.50$ for Subtask 1 and Subtask 2, respectively, compared with $73.63/76.51$ for Expert-3Y and $81.83/80.07$ for Expert-8Y.
For Subtask 1, the best AI (SJTU1) significantly outperformed Expert-3Y ($p=0.003$, $q=0.044$ after Benjamini–Hochberg correction).
A similar pattern was observed in the non-contrast setting, where the best AI achieved AUCs of $83.76$ and $81.75$ for Subtask 1 and Subtask 2, respectively, compared with $69.53/69.76$ for Expert-3Y and $77.29/74.96$ for Expert-8Y, again placing the best AI around the senior-radiologist reference and above the junior-radiologist reference.
For Subtask 1, the best AI (SDU) significantly outperformed Expert-3Y ($p=0.001$, $q=0.043$ after Benjamini–Hochberg correction), and for Subtask 2 the best AI (UCL) exceeded Expert-3Y at the unadjusted level only ($p=0.017$, $q=0.158$).
Taken together, these results show that the best-performing AI methods can significantly outperform the junior radiologist on selected endpoints while approaching the senior-radiologist reference.
Median AI performance was more consistent with a junior-reader benchmark: it exceeded Expert-3Y across all four ID cells, yet did not differ significantly from either reader in any cell after multiplicity correction.

\begin{table}[!t]
\centering
\caption{
Inter-reader agreement between Expert-3Y and Expert-8Y.
$\kappa_{w}$ denotes quadratic-weighted Cohen's $\kappa$, and $\kappa$ denotes unweighted Cohen's $\kappa$ for the binarised Subtask~1.
Bracketed ranges indicate $95\%$ bootstrap confidence intervals.}
\label{tab:reader_reference}
\setlength{\tabcolsep}{4pt}\renewcommand{\arraystretch}{1.15}
\resizebox{\columnwidth}{!}{
\begin{tabular}{llcc}
\toprule
\textbf{Domain} & \textbf{Setting} & \textbf{$\kappa_{w}$ (7-point)} & \textbf{$\kappa$ (Subtask 1)} \\
\midrule
ID  & Non-contrast      & 0.625 [0.517, 0.709] & 0.475 [0.324, 0.615] \\
ID  & Contrast-enhanced & 0.661 [0.562, 0.739] & 0.458 [0.318, 0.592] \\
OOD & Non-contrast      & 0.439 [0.222, 0.608] & 0.260 [0.076, 0.449] \\
OOD & Contrast-enhanced & 0.318 [0.111, 0.522] & 0.195 [0.072, 0.330] \\
\bottomrule
\end{tabular}}
\end{table}

\textbf{Under OOD evaluation.}
In the contrast-enhanced setting, the best AI achieved AUCs of $77.98$ and $79.69$, numerically exceeding the senior-radiologist reference for both subtasks, although neither difference reached statistical significance, and also outperforming the junior-radiologist reference.
In the non-contrast setting, the best AI achieved the highest AUC for Subtask 2 ($86.31$), whereas the senior radiologist achieved the highest AUC for Subtask 1 ($77.57$), exceeding the best AI result of $71.51$.
The median AI performance also decreased under OOD evaluation, from $78.31$ to $61.05$ in contrast-enhanced Subtask 1, from $78.45$ to $52.62$ in contrast-enhanced Subtask 2, from $79.78$ to $58.74$ in non-contrast Subtask 1, and from $74.28$ to $60.00$ in non-contrast Subtask 2.
These findings indicate that the favorable clinical positioning observed under ID evaluation does not persist consistently under distribution shift.

\textbf{Uncertainty in clinical comparison.}
Table~\ref{tab:results_merged} reports two-sided $95\%$ DeLong confidence intervals for all AUC estimates together with the corresponding positive and negative case counts.
Under ID evaluation, all AUC confidence intervals exclude chance ($50\%$), with median widths of $16.9$ and $17.7$ AUC points for Subtasks~1 and~2, respectively.
Under OOD evaluation, the median interval widths increase to $28.7$ and $51.5$ points, with $14$ of $22$ Subtask~1 AUCs and $16$ of $22$ Subtask~2 AUCs having confidence intervals that include chance.
Uncertainty is particularly pronounced for OOD Subtask~2, which contains only $5$ positive and $65$ negative cases, $10$ of $22$ accuracies equal the $92.86\%$ majority-class rate, while the radiologist AUC confidence intervals span $23$--$40$ points.
We therefore treat OOD Subtask~2 as descriptive and do not rank methods on this endpoint.
The clinical reference also showed greater inter-reader variability under OOD evaluation: quadratic-weighted $\kappa$ decreased from $0.625/0.661$ for non-contrast/contrast-enhanced ID readings to $0.439/0.318$ for the corresponding OOD readings (Table~\ref{tab:reader_reference}).

Overall, these results position the current AI frontier near experienced-radiologist performance in selected settings, with significant gains over the junior radiologist on selected ID endpoints, while showing that this clinical proximity remains conditional on acquisition setting, domain, and diagnostic endpoint.

\begin{figure*}[t!]
    \centering
    \begin{minipage}[t]{0.44\textwidth}
        \centering
        \includegraphics[width=\linewidth]{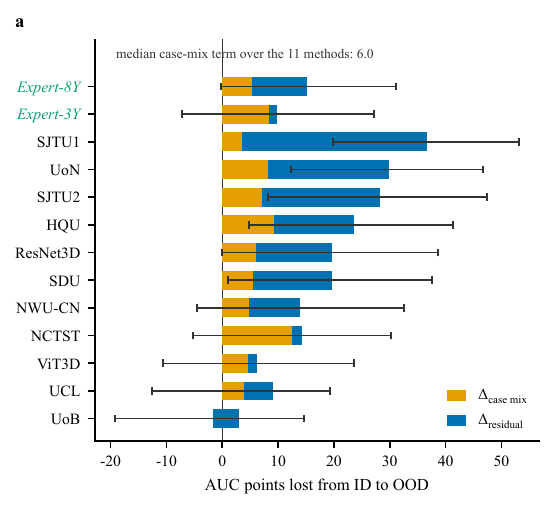}
    \end{minipage}\hfill
    \begin{minipage}[t]{0.54\textwidth}
        \centering
        \includegraphics[width=\linewidth]{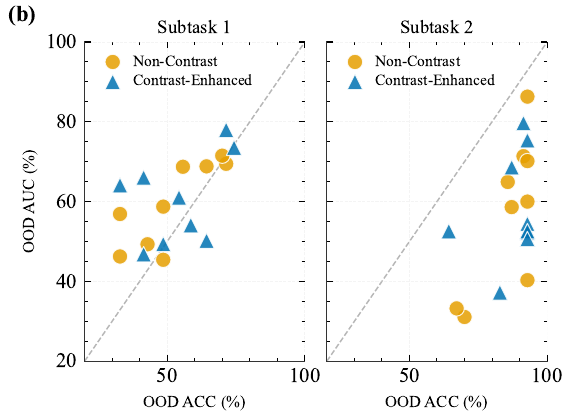}
    \end{minipage}

    \vspace{0.6em}

    \begin{minipage}[t]{0.44\textwidth}
        \centering
        \includegraphics[width=\linewidth]{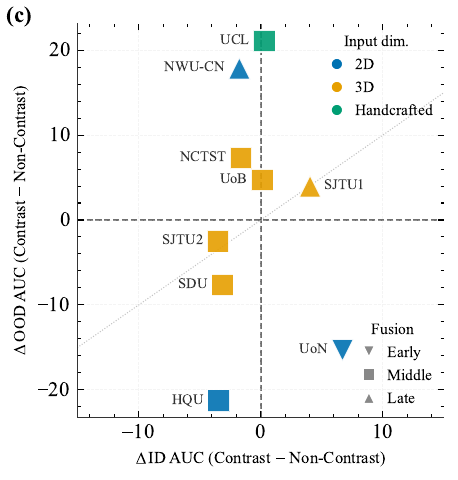}
    \end{minipage}\hfill
    \begin{minipage}[t]{0.54\textwidth}
        \centering
        \includegraphics[width=\linewidth]{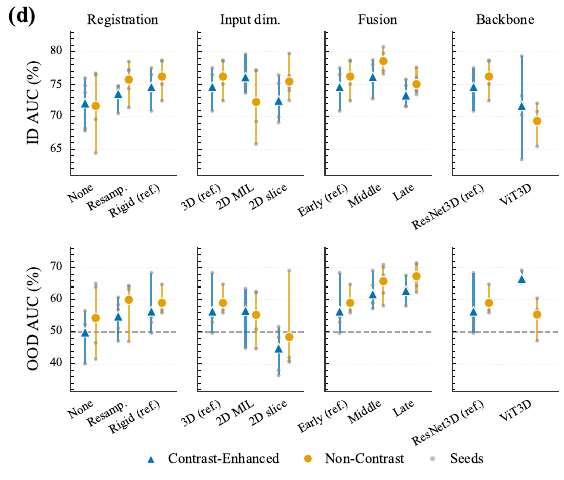}
    \end{minipage}

    \caption{Comprehensive analysis of challenge results across multiple dimensions.
(a) Stage-standardised decomposition of the contrast-enhanced Subtask~1 ID--OOD AUC loss into fibrosis-stage composition and residual components, with $95\%$ stratified-bootstrap confidence intervals for the residual. Radiologists are italicised; positive values indicate lower OOD AUC.
(b) ACC--AUC discrepancy illustrates label-imbalance effects in Subtask~2.
(c) Contrast-enhanced sequences show method-dependent effects under OOD conditions. HQU is grouped under 2D for visualization.
(d) Controlled one-factor-at-a-time analysis of the same four design choices in Subtask~1, using five seeds per arm. Markers show seed means, bars the observed seed ranges, grey dots individual seeds, and ``ref.'' the anchor configuration. Backbone values are seed means and therefore differ from the five-seed probability ensembles reported in Table~\ref{tab:results_merged}.}
    \label{fig:combined}
\end{figure*}

\subsection{Data-perspective challenges in real-world deployment}\label{subsec:data_challenges}

To investigate what drives AI's loss of clinically meaningful performance under realistic conditions, we examined three data-side factors that emerged from the LiFS benchmark: cross-center distribution shift, label imbalance in evaluation, and the variable behavior of contrast-enhanced sequences.

\textbf{OOD Generalization.}
Vendor C changes center, scanner, field strength, protocol, and fibrosis-stage composition simultaneously, so the unadjusted ID--OOD difference cannot identify which factor caused the change. We therefore decomposed the AUC drop after standardising the ID cohort to the OOD stage composition. 
In the primary contrast-enhanced Subtask~1 analysis, the median AUC drop across the 11 AI methods was $19.6$ AUC points, comprising a median $6.0$-point case-mix component and a median $13.7$-point residual non-stage component (Fig.~\ref{fig:combined}(a)). 
Residuals ranged from $-5.3$ to $33.3$ points, and only SJTU1, UoN, and SJTU2 had residual confidence intervals excluding zero. 
The wide method-to-method spread therefore persisted after adjustment, but most individual estimates remained uncertain.

An exact stage-matched sensitivity analysis closely agreed with the weighted estimates: across all $52$ method/reader cells, the maximum deviation was $0.195$ bootstrap SD, although matching retained only $17$ ID negatives (S1:4, S2:7, S3:6).
Subtracting each radiologist's ID--OOD loss gave a reader-anchored excess loss for the median method of $9.8$, $-0.1$, $4.3$ and $13.5$ AUC points against Expert-3Y across the four evaluation cells, and $4.4$, $15.3$, $19.3$ and $22.9$ points against Expert-8Y: positive in seven of the eight comparisons, but anchor-dependent, because the readers' own losses were not case-mix alone (residual components $-5.2$ to $23.5$ points, including $23.5$ of Expert-3Y's $24.0$-point contrast-enhanced Subtask~2 loss, 95\% CI $4.1$--$40.1$). 
This shift is therefore visible to human readers as well.
The decomposition applied to all $100$ controlled runs of Section~\ref{subsec:design_choices} independently reproduced this pattern, with median total, case-mix and residual components of $17.8$, $5.6$ and $12.6$ AUC points.

Vendor~C showed near-complete sequence availability and image-derived SNR comparable to Vendor~B1. At the same time, models varied by a median absolute $12.2$ AUC points between Vendors~A and B1, two 3.0-T scanners at the same center, compared with $5.5$ and $5.8$ points for the two radiologists. Together, these descriptive findings are consistent with limited model transfer across acquisition environments, including between scanners at the same center and field strength. Center, scanner model, field strength, protocol, and unmeasured clinical differences remain inseparable in the present design.

\begin{table*}[!t]
\centering
\caption{Run-level controlled factor analysis on Subtask~1. Each block varies one factor while holding all others at the reference setting (\colorbox{black!8}{shaded}). CE and NC AUCs are reported as mean $\pm$ sample SD across five seeds. For each non-reference level, $\Delta_{\mathrm{seed}}$ is the mean seed-paired arm-minus-reference effect, averaged across CE and NC within each seed. Two-sided $95\%$ CIs and $p$-values are based on one-sample $t$ inference over the five seed-level effects; $q$-values use Benjamini–Hochberg correction within seven prespecified ID and seven OOD contrasts. $^{\ast}$ denotes $q<0.05$ and $^{\dagger}$ denotes nominal significance only ($p<0.05$, $q\geq0.05$); reference rows have no inferential statistics. GFLOPs are reported for CE/NC. Reference values are seed means and therefore differ from the five-seed probability ensembles in Table~\ref{tab:results_merged}.}
\label{tab:controlled_factors}
\setlength{\tabcolsep}{3pt}\renewcommand{\arraystretch}{1.15}
\resizebox{\textwidth}{!}{\small
\renewcommand{\cellset}{\renewcommand{\arraystretch}{0.85}}
\newcolumntype{A}{>{\centering\arraybackslash}m{46pt}}
\newcolumntype{D}{>{\centering\arraybackslash}m{92pt}}
\newcolumntype{F}{>{\raggedright\arraybackslash}m{62pt}}
\newcolumntype{V}{>{\raggedright\arraybackslash}m{54pt}}
\newcolumntype{G}{>{\centering\arraybackslash}m{50pt}}
\begin{tabular}{F V ?AA?AA?DD?G}
\toprule
\multirow{2}{*}{\textbf{Factor}} & \multirow{2}{*}{\textbf{Level}}
& \multicolumn{2}{c}{\cellcolor{cebg}\textbf{Contrast-Enhanced (CE)}}
& \multicolumn{2}{c}{\cellcolor{ncbg}\textbf{Non-Contrast (NC)}}
& \multicolumn{2}{c}{\cellcolor{hdrbg}\textbf{Run-level effect across CE/NC}}
& \multirow{2}{*}{\makecell{\textbf{GFLOPs}\\[1pt]\tabci{CE\,/\,NC}}} \\
\cmidrule(lr){3-4}\cmidrule(lr){5-6}\cmidrule(lr){7-8}
& & \makecell{\textbf{ID}\\\textbf{AUC}} & \makecell{\textbf{OOD}\\\textbf{AUC}}
  & \makecell{\textbf{ID}\\\textbf{AUC}} & \makecell{\textbf{OOD}\\\textbf{AUC}}
  & \makecell{$\bm{\Delta}_{\mathrm{seed}}$\ \textbf{ID}\\\tabci{$95\%$ CI; $p,q$}}
  & \makecell{$\bm{\Delta}_{\mathrm{seed}}$\ \textbf{OOD}\\\tabci{$95\%$ CI; $p,q$}} & \\
\midrule
\rowcolor{black!8}
\textbf{Registration} & Rigid
& \makecell{$0.746$\\[1pt]\tabci{$\pm0.026$}} & \makecell{$0.564$\\[1pt]\tabci{$\pm0.072$}}
& \makecell{$0.762$\\[1pt]\tabci{$\pm0.026$}} & \makecell{$0.590$\\[1pt]\tabci{$\pm0.035$}}
& \refcell & \refcell
& $29.3$\,/\,$20.3$ \\
 & None
& \makecell{$0.721$\\[1pt]\tabci{$\pm0.038$}} & \makecell{$0.499$\\[1pt]\tabci{$\pm0.061$}}
& \makecell{$0.717$\\[1pt]\tabci{$\pm0.051$}} & \makecell{$0.543$\\[1pt]\tabci{$\pm0.104$}}
& \makecell{$-0.035$\\[1pt]\tabci{$[-0.081,+0.011]$}\\[0pt]\tabci{$p{=}0.100,\ q{=}0.140$}} & \makecell{$-0.056$\\[1pt]\tabci{$[-0.178,+0.066]$}\\[0pt]\tabci{$p{=}0.272,\ q{=}0.381$}}
& $29.3$\,/\,$20.3$ \\
 & Resample
& \makecell{$0.736$\\[1pt]\tabci{$\pm0.018$}} & \makecell{$0.549$\\[1pt]\tabci{$\pm0.056$}}
& \makecell{$0.757$\\[1pt]\tabci{$\pm0.028$}} & \makecell{$0.600$\\[1pt]\tabci{$\pm0.074$}}
& \makecell{$-0.008$\\[1pt]\tabci{$[-0.025,+0.009]$}\\[0pt]\tabci{$p{=}0.276,\ q{=}0.276$}} & \makecell{$-0.003$\\[1pt]\tabci{$[-0.100,+0.095]$}\\[0pt]\tabci{$p{=}0.941,\ q{=}0.941$}}
& $29.3$\,/\,$20.3$ \\
\addlinespace[2.5pt]
\midrule
\rowcolor{black!8}
\textbf{Input dim.} & 3D volume
& \makecell{$0.746$\\[1pt]\tabci{$\pm0.026$}} & \makecell{$0.564$\\[1pt]\tabci{$\pm0.072$}}
& \makecell{$0.762$\\[1pt]\tabci{$\pm0.026$}} & \makecell{$0.590$\\[1pt]\tabci{$\pm0.035$}}
& \refcell & \refcell
& $29.3$\,/\,$20.3$ \\
 & 2D MIL
& \makecell{$0.761$\\[1pt]\tabci{$\pm0.024$}} & \makecell{$0.566$\\[1pt]\tabci{$\pm0.069$}}
& \makecell{$0.723$\\[1pt]\tabci{$\pm0.049$}} & \makecell{$0.553$\\[1pt]\tabci{$\pm0.076$}}
& \makecell{$-0.012$\\[1pt]\tabci{$[-0.035,+0.011]$}\\[0pt]\tabci{$p{=}0.218,\ q{=}0.255$}} & \makecell{$-0.018$\\[1pt]\tabci{$[-0.082,+0.047]$}\\[0pt]\tabci{$p{=}0.486,\ q{=}0.567$}}
& $69.3$\,/\,$66.7$ \\
 & \makecell[l]{2D single\\slice}
& \makecell{$0.725$\\[1pt]\tabci{$\pm0.032$}} & \makecell{$0.449$\\[1pt]\tabci{$\pm0.071$}}
& \makecell{$0.754$\\[1pt]\tabci{$\pm0.028$}} & \makecell{$0.484$\\[1pt]\tabci{$\pm0.121$}}
& \makecell{$-0.014$\\[1pt]\tabci{$[-0.033,+0.004]$}\\[0pt]\tabci{$p{=}0.095,\ q{=}0.140$}} & \makecell{$-0.111$\\[1pt]\tabci{$[-0.248,+0.027]$}\\[0pt]\tabci{$p{=}0.089,\ q{=}0.275$}}
& $2.2$\,/\,$2.1$ \\
\addlinespace[2.5pt]
\midrule
\rowcolor{black!8}
\textbf{Fusion depth} & Early
& \makecell{$0.746$\\[1pt]\tabci{$\pm0.026$}} & \makecell{$0.564$\\[1pt]\tabci{$\pm0.072$}}
& \makecell{$0.762$\\[1pt]\tabci{$\pm0.026$}} & \makecell{$0.590$\\[1pt]\tabci{$\pm0.035$}}
& \refcell & \refcell
& $29.3$\,/\,$20.3$ \\
 & Middle
& \makecell{$0.762$\\[1pt]\tabci{$\pm0.024$}} & \makecell{$0.618$\\[1pt]\tabci{$\pm0.049$}}
& \makecell{$0.786$\\[1pt]\tabci{$\pm0.017$}} & \makecell{$0.658$\\[1pt]\tabci{$\pm0.055$}}
& \makecell{$+0.020^{\dagger}$\\[1pt]\tabci{$[+0.002,+0.037]$}\\[0pt]\tabci{$p{=}0.038,\ q{=}0.132$}} & \makecell{$+0.061$\\[1pt]\tabci{$[-0.024,+0.146]$}\\[0pt]\tabci{$p{=}0.118,\ q{=}0.275$}}
& $90.8$\,/\,$41.1$ \\
 & Late
& \makecell{$0.733$\\[1pt]\tabci{$\pm0.019$}} & \makecell{$0.628$\\[1pt]\tabci{$\pm0.034$}}
& \makecell{$0.750$\\[1pt]\tabci{$\pm0.017$}} & \makecell{$0.674$\\[1pt]\tabci{$\pm0.040$}}
& \makecell{$-0.012$\\[1pt]\tabci{$[-0.028,+0.003]$}\\[0pt]\tabci{$p{=}0.091,\ q{=}0.140$}} & \makecell{$+0.074^{\dagger}$\\[1pt]\tabci{$[+0.032,+0.115]$}\\[0pt]\tabci{$p{=}0.008,\ q{=}0.054$}}
& $110.8$\,/\,$47.5$ \\
\addlinespace[2.5pt]
\midrule
\rowcolor{black!8}
\textbf{Backbone} & \makecell[l]{ResNet3D\\\tabci{(CNN)}}
& \makecell{$0.746$\\[1pt]\tabci{$\pm0.026$}} & \makecell{$0.564$\\[1pt]\tabci{$\pm0.072$}}
& \makecell{$0.762$\\[1pt]\tabci{$\pm0.026$}} & \makecell{$0.590$\\[1pt]\tabci{$\pm0.035$}}
& \refcell & \refcell
& $29.3$\,/\,$20.3$ \\
 & \makecell[l]{ViT3D\\\tabci{(Transformer)}}
& \makecell{$0.717$\\[1pt]\tabci{$\pm0.057$}} & \makecell{$0.666$\\[1pt]\tabci{$\pm0.016$}}
& \makecell{$0.693$\\[1pt]\tabci{$\pm0.025$}} & \makecell{$0.554$\\[1pt]\tabci{$\pm0.049$}}
& \makecell{$-0.049^{\dagger}$\\[1pt]\tabci{$[-0.080,-0.018]$}\\[0pt]\tabci{$p{=}0.012,\ q{=}0.081$}} & \makecell{$+0.033$\\[1pt]\tabci{$[-0.032,+0.097]$}\\[0pt]\tabci{$p{=}0.231,\ q{=}0.381$}}
& $5.0$\,/\,$3.8$ \\
\bottomrule
\end{tabular}}
\end{table*}

\textbf{Label Imbalance.}
Fig.~\ref{fig:combined}(b) plots the relationship between ACC and AUC metrics for all methods on the OOD test set.
The analysis reveals a notable discrepancy between ACC and AUC metrics for Subtask 2, where several methods achieved ACC exceeding $85\%$ while their AUC remained below $80\%$.
In contrast, Subtask 1 shows relatively stronger ACC-AUC agreement, whereas Subtask 2 exhibits a more evident discrepancy between high ACC and modest or low AUC.
This pattern is consistent with the severe majority-class dominance in Subtask 2, where S2--S4 cases constitute approximately $92.9\%$ of the OOD test set, shown in Fig.~\ref{fig:vendor_partition}(c).
The recurring ACC value of $92.86\%$ across multiple methods in Subtask 2, particularly paired with low AUCs, strongly indicates model collapse, where algorithms default to predicting the majority class.
This confirms that ACC is an unreliable metric for this benchmark due to the skewed distribution~\citep{J_NM_metric}.
These findings underscore the importance of reporting multiple evaluation metrics, particularly AUC, which provides a more reliable assessment of model discriminative ability when class distributions are imbalanced.

\textbf{Contrast-Enhanced Sequences.}
We further explored the impact of including contrast-enhanced sequences on liver fibrosis staging performance.
Fig.~\ref{fig:combined}(c) illustrates the performance change from Non-Contrast to Contrast-Enhanced settings for each team under both ID and OOD conditions for Subtask 1 (S4 vs. S1--3).
Under ID conditions, the inclusion of GED sequences yielded only marginal performance changes, with AUC fluctuations remaining below $5\%$ for the majority of teams and the largest improvement being a modest $6.7\%$ (Team UoN).
This suggests that standard non-contrast sequences (T1W, T2W, DWI) already provide sufficient discriminatory features when training and test distributions are well-aligned.
Moreover, several teams (SJTU2, HQU, SDU, NWU-CN, NCTST) experienced performance degradation, which may reflect a form of hidden distribution shift intrinsic to the contrast-enhanced phases themselves, especially the HBP.
Under OOD conditions, however, the effect of contrast enhancement becomes far more pronounced, yet in opposite directions for different methods.
Team UCL and NWU-CN achieved substantial AUC improvements of $21.1\%$ and $17.9\%$, respectively, consistent with the hypothesis that HBP signals capture functional liver information that is more invariant to scanner-induced shifts than anatomical texture.
In stark contrast, Team HQU and Team UoN suffered severe OOD degradation of $-21.3\%$ and $-15.4\%$, respectively.
Overall, these results suggest that contrast-enhanced imaging, especially HBP, provides valuable functional information, but its performance impact varies across methods and evaluation settings, highlighting a practical dilemma in its use under cross-center evaluation.

\subsection{Technical-perspective challenges}
\label{subsec:design_choices}

The evaluated methods differ in several design choices simultaneously, so comparisons across them cannot isolate the effect of any single factor.
To ask which design choices actually matter, we ran a controlled one-factor-at-a-time study in which each factor is varied in isolation with the rest of the pipeline held identical (Fig.~\ref{fig:combined}(d)).

\textbf{Setting.}
All controlled factor contrasts target Subtask~1 using the same challenge split of $360$ training, $60$ validation, and $190$ test cases, with the test set comprising $120$ ID and $70$ OOD cases.
Experiments are conducted separately under the non-contrast setting (T1W, T2W, and DWI) and the contrast-enhanced setting (T1W, T2W, DWI, and GED1--4).
Each sequence is normalized using case-wise non-zero-voxel z-scoring, and the full volumes are resized or resampled to $32\times160\times160$ voxels.
We use a strict one-factor-at-a-time design anchored on a common reference configuration consisting of rigid registration, 3D input, early fusion, and ResNet3D.
The four controlled factors and their evaluated levels are registration (none, resampling, and rigid registration), input dimensionality (single-slice 2D, all-slice 2D multiple instance learning (MIL) with mean pooling, and 3D), fusion strategy (early, middle, and late fusion), and backbone architecture (ResNet3D and ViT3D).
For each comparison, only the factor of interest is changed from the reference configuration, while all other settings are held fixed.
Middle and late fusion use shared-weight sequence branches, with fusion performed at the intermediate feature-map and pooled feature-vector levels, respectively.

All models are trained from scratch using cross-entropy loss, AdamW with a learning rate of $2\times10^{-4}$ and weight decay of $10^{-4}$, cosine learning-rate scheduling, $100$ epochs, batch size $16$, and the same training augmentation.
Each reported configuration is trained with five matched seeds.
Model selection is based only on validation AUC, and the test set is evaluated once per run.
For each non-reference level, acquisition track, domain, and seed, the run-level effect is the paired AUC difference from the reference arm. 
Because CE and NC share the same seeds and case partition, their paired differences were averaged within each seed, yielding five independent seed-level effects for the primary one-sample $t$ inference. 
Benjamini–Hochberg adjustment was applied separately to the prespecified families of seven ID and seven OOD factor-level hypotheses. Table~\ref{tab:controlled_factors} reports the resulting effect estimates, $95\%$ confidence intervals, $p$-values, and $q$-values. 

\textbf{Overall findings.}
Across the five reference runs, the standard deviation in AUC was $0.026$ under ID evaluation for both acquisition settings and ranged from $0.035$ to $0.072$ under OOD evaluation, providing a reference scale for interpreting the factor effects below.
Overall, most individual factor changes produced modest effects, with uncertainty generally large relative to the observed differences.
At the run level, three of the $14$ factor–domain contrasts reached nominal $p<0.05$, but none remained below $q=0.05$ after Benjamini–Hochberg correction. The smallest adjusted value was observed for late-fusion OOD AUC ($q=0.054$).
A complementary case-level analysis of the five-seed averaged predictions yielded similarly uncertain estimates: all $28$ per-track domain-specific bootstrap $95\%$ confidence intervals included zero, and all paired DeLong $p$-values exceeded $0.05$.
These results reflect substantial case-sampling uncertainty in the estimated differences between the seed-averaged predictors.
The OOD split is also substantially imbalanced, with a majority-class accuracy of $0.671$. We therefore focus on AUC when interpreting OOD factor effects.
The following analyses examine the magnitude and direction of each factor effect under both acquisition settings.

\textbf{Pre-registration.}
Omitting registration yielded the lowest mean AUC among the three levels in all four evaluation cells.
On the contrast-enhanced track, ID AUC increased from $0.721$ to $0.736$ to $0.746$, and OOD AUC from $0.499$ to $0.549$ to $0.564$, across no registration, shared-grid resampling, and rigid registration, respectively.
Decomposing the pipeline into its two steps localises the more consistent OOD gain to the shared grid: resampling improved OOD AUC by $+0.050$ on CE and $+0.057$ on NC, whereas adding Euler3D rigid registration contributed only $+0.015$ and $-0.010$, respectively.
Across tracks, omitting registration reduced mean ID AUC by $0.035$ (95\% CI $[-0.081,+0.011]$, $p=0.100$, $q=0.140$).
Shared-grid resampling versus rigid registration was $-0.008$ (95\% CI $[-0.025,+0.009]$, $p=0.276$, $q=0.276$).
The descriptive ordering suggests that placing sequences on a shared physical grid may contribute more consistently to OOD performance than additional rigid motion correction.

\textbf{Input dimensionality.}
The five-seed analysis separates the two 2D formulations.
For 2D MIL, the across-track ID effect was $-0.012$ (95\% CI $[-0.035,+0.011]$, $p=0.218$,
$q=0.255$), while its OOD interval was wider. Reducing the volume to a single central slice had an
ID effect of $-0.014$ (95\% CI $[-0.033,+0.004]$, $p=0.095$, $q=0.140$) and an OOD effect of
$-0.111$ (95\% CI $[-0.248,+0.027]$, $p=0.089$, $q=0.275$), with similar mean seed-paired
effects on CE ($-0.115$) and NC ($-0.106$).
The single-slice model also remained variable, spanning an OOD AUC of $0.407$--$0.691$ across the five non-contrast seeds.
The computational trade-off is more pronounced, as shown in Table~\ref{tab:controlled_factors}.
The 2D single-slice model requires $2.2/2.1$ GFLOPs compared with $29.3/20.3$ GFLOPs for the 3D volume on CE/NC, a $9.8$--$13.5\times$ reduction.
Thus, single-slice 2D substantially reduces computation while showing a consistent directional OOD penalty across tracks.

\textbf{Fusion strategy.}
Fusion shows the most directionally consistent pattern among the controlled factors.
Moving fusion later improves OOD AUC in all four per-track comparisons, with gains of $+0.054$ to $+0.083$ (mean $+0.067$), and reduces the ID--OOD gap in all four ($+0.038$ to $+0.095$).
Accordingly, early fusion yields the lowest OOD AUC on both tracks ($0.564$ CE and $0.590$ NC), compared with $0.618/0.658$ for middle fusion and $0.628/0.674$ for late fusion.
Across tracks, middle fusion changed OOD AUC by $+0.061$ (95\% CI $[-0.024,+0.146]$,
$p=0.118$, $q=0.275$), and late fusion changed it by $+0.074$ (95\% CI
$[+0.032,+0.115]$, $p=0.008$, $q=0.054$). Middle fusion's ID gain reached the nominal threshold
but not the adjusted threshold ($+0.020$, 95\% CI $[+0.002,+0.037]$, $p=0.038$, $q=0.132$),
whereas late fusion's ID effect was $-0.012$ (95\% CI $[-0.028,+0.003]$, $p=0.091$,
$q=0.140$).
These OOD gains come at a substantial computational cost: late fusion requires $110.8/47.5$ GFLOPs compared with $29.3/20.3$ GFLOPs for early fusion on CE/NC, or $3.8\times/2.3\times$ as much computation.

This OOD fusion benefit appears to be conditional on registration rather than intrinsic to the
fusion depth. In a balanced $2\times3$ registration-by-fusion block, run with the same five seed
clusters as the primary analysis, OOD AUC rises across the fusion ladder when the sequences are
rigidly registered ($0.564\rightarrow0.618\rightarrow0.628$ on CE and
$0.590\rightarrow0.658\rightarrow0.674$ on NC), whereas without registration the same ladder is
flat or decreasing ($0.499\rightarrow0.475\rightarrow0.492$ on CE and
$0.543\rightarrow0.520\rightarrow0.489$ on NC). The corresponding track-averaged OOD interaction
contrasts are $-0.084$ for middle versus early fusion (95\% CI $[-0.224,+0.055]$, $p=0.168$) and
$-0.104$ for late versus early fusion (95\% CI $[-0.259,+0.050]$, $p=0.134$); the ID interaction
terms are similarly imprecise. 
The descriptive pattern suggests that the OOD benefit of later fusion may depend on sequences sharing a common physical grid.

Overall, later fusion trades substantially greater computation for directionally higher OOD AUC.

\textbf{Backbone architecture.}
The effect of backbone architecture varies substantially across acquisition settings.
Under ID evaluation, ViT3D performs below the ResNet3D reference on both tracks, with mean
seed-paired AUC differences of $-0.029$ for contrast-enhanced and $-0.068$ for non-contrast input;
the across-track effect was $-0.049$ (95\% CI $[-0.080,-0.018]$, $p=0.012$, $q=0.081$).
Under OOD evaluation, however, ViT3D improves mean AUC by $+0.102$ on the contrast-enhanced track,
while the effect reverses to $-0.036$ on the non-contrast track, leaving the across-track effect
imprecise ($+0.033$, 95\% CI $[-0.032,+0.097]$, $p=0.231$, $q=0.381$).
Thus, the OOD effect of backbone architecture is strongly track-dependent, with ViT3D performing better on contrast-enhanced input but worse on non-contrast input.
All four per-track case-level bootstrap intervals span zero.
ViT3D also requires substantially less computation, at $5.0/3.8$ GFLOPs compared with $29.3/20.3$ GFLOPs for ResNet3D on CE/NC.
Overall, the relative performance of ViT3D and ResNet3D varies substantially with the available imaging sequences.

\textbf{Summary.}
Across the controlled analyses, later fusion showed the most consistent directional OOD gain, single-slice input showed a similar OOD penalty across tracks, and the ViT3D OOD effect reversed between contrast-enhanced and non-contrast input. No factor–domain contrast survived the prespecified multiplicity correction when CE and NC were clustered within the five shared seeds. Seed-level and case-level analyses quantify complementary sources of uncertainty.

\section{Discussion and Conclusion}

This LiFS benchmark was designed to answer a central question: how far has current AI progressed toward clinically meaningful performance in liver fibrosis staging from real-world MRI? 
The results suggest that AI has made meaningful progress, but that clinically meaningful performance remains conditional. 
Under acquisition conditions similar to the training data, leading AI methods can approach the performance range of experienced radiologists, and in selected exploratory comparisons they exceeded the junior radiologist. 
This performance pattern suggests that MRI-based AI has the potential to support non-invasive fibrosis staging and, with further prospective validation, may help reduce reliance on biopsy in selected clinical scenarios.
However, the same level of performance was not consistently maintained across centers, acquisition settings, and diagnostic subtasks. 
Median AI performance was generally less competitive, and cross-center evaluation exposed substantial instability. 
Therefore, the current frontier is not simply achieving high performance on matched test data, but establishing whether learned fibrosis-related imaging patterns remain reliable under heterogeneous clinical acquisition.

The comparison with radiologists helps define this gap in clinically interpretable terms. 
The greater cross-center instability of AI relative to the radiologists shows that strong in-distribution performance alone is insufficient evidence for robust clinical performance.

The stage-standardised analysis further shows that differences in fibrosis-stage composition explain only part of the cross-center performance loss, with substantial method-dependent degradation remaining after adjustment. This residual gap is consistent with limited transfer of learned imaging features across heterogeneous acquisition environments. 

Beyond the AI versus radiologist comparison, LiFS also identifies the major data-side and technical challenges that must be addressed for reliable liver fibrosis staging. On the data side, these include cross-center acquisition heterogeneity, label imbalance, and the variable contribution of contrast-enhanced sequences, especially the HBP.
On the technical side, the benchmark highlights the need for more robust spatial alignment, multi-modal fusion, missing-modality handling, and architecture design under domain shift. Because the evaluated methods differ in several coupled design choices at once, we examined these factors directly in a controlled one-factor-at-a-time study (Section~\ref{subsec:design_choices}). The five-seed analysis showed directional OOD penalties for single-slice input and gains for later fusion, but no factor--domain contrast survived multiplicity correction after CE and NC were clustered within their shared seeds; late-fusion OOD AUC was closest ($q=0.054$). The separate patient-level analysis of seed-averaged predictions also produced wide intervals. Design guidance derived from a leaderboard therefore needs paired, seed-repeated contrasts under a fixed pipeline, larger patient samples, and prospective multi-center validation.
\paragraph{Limitations}
The current cohort of 610 patients from three centers also represents a limited spectrum of clinical MRI practice, and broader multi-center validation will be important to further assess the generalizability of the benchmark findings.
The multiple factors confounded in Vendor~C cannot be disentangled, precluding attribution of the residual gap to any individual source of distribution shift.
The clinical reference is also limited to two individual in-cohort readers without consensus arbitration. The AI-versus-radiologist comparison is therefore exploratory, and broader validation with a larger reader panel and longer inter-session washout is needed to assess how well these findings extend to radiologist-level performance more generally.

Overall, LiFS shows that MRI-based AI can approach a clinically meaningful level in selected in-distribution settings, suggesting its potential to support non-invasive liver fibrosis staging and, with further prospective validation, to help reduce reliance on biopsy in selected clinical scenarios. 
At the same time, its robustness remains insufficient for reliable deployment across heterogeneous clinical environments. By combining multi-center MRI, complete gadoxetic acid-enhanced sequences, histopathology-confirmed labels, in-cohort radiologist references, and independently developed methods, LiFS provides a realistic benchmark for assessing how far AI has progressed in liver fibrosis staging. Importantly, it exposes the gap between strong benchmark performance and clinically reliable AI, and offers a unified platform for studying the key challenges that still limit deployment, including cross-center generalization, imbalance-aware evaluation, robust use of contrast-enhanced imaging, and uncertainty-aware multi-modal modeling. In this sense, LiFS aims to measure current progress while guiding the next stage of research toward reliable, clinically deployable AI for liver fibrosis assessment.

\section*{Acknowledgement}

This work was funded by the Science and Technology Commission of Shanghai Municipality (25TS1412100), the Noncommunicable Chronic Disease-National Science and Technology Major Project (2026ZD0555800/2026ZD0555802), and the National Natural Science Foundation of China (62372115).

\section*{Declaration of competing interest}

The authors declare that they have no known competing financial interests or personal relationships that could have appeared to influence the work reported in this paper (Conflicts of interest: none).

\section*{Declaration of generative AI and AI-assisted technologies in the writing process}

During the preparation of this work, the authors used ChatGPT to assist with language polishing, grammar refinement, and clarity improvement. The tool was not used to generate scientific results, perform experiments, create data, fabricate citations, or make unsupported scientific claims. After using this tool, the authors reviewed and edited the content as needed and take full responsibility for the content of the published article.

\bibliographystyle{model2-names}\biboptions{authoryear}
\bibliography{strings,refs}

\clearpage
\end{document}